\documentclass[10pt,twocolumn,letterpaper]{article}

\usepackage[pagenumbers]{wacv} 

\usepackage{newtxtext}
\usepackage{enumitem} 

\usepackage{float}
\usepackage{listings}
\usepackage{fancybox}

\usepackage{multirow}
\usepackage{algorithm}
\usepackage{algorithmic}
\usepackage{threeparttable}
\usepackage{cuted}
\usepackage{wrapfig}
\usepackage{xspace}
\usepackage{amsmath}
\usepackage{amsthm}
\usepackage{subcaption}
\usepackage{tabularx}
\usepackage{booktabs}
\usepackage{caption}
\usepackage{colortbl}
\usepackage{xcolor}
\usepackage{comment}

\newtheorem{theorem}{Theorem}[section]
\newtheorem{corollary}{Corollary}[theorem]

\newtheorem{remark}{Remark} 
\theoremstyle{definition}

\usepackage{pgfplots}
\pgfplotsset{compat=1.18} 

\def\convnn{\textsc{ConvNN}\xspace}

\def\argkmax{\text{$k$-argmax}}
\def\kmax{\text{$k$-max}}

\definecolor{vsc-bg}{HTML}{FFFFFF}
\definecolor{vsc-keyword}{HTML}{0000FF}
\definecolor{vsc-identifier}{HTML}{001080}
\definecolor{vsc-string}{HTML}{A31515}
\definecolor{vsc-comment}{HTML}{008000}
\definecolor{vsc-number}{HTML}{098658}
\definecolor{vsc-builtin}{HTML}{795E26}
\definecolor{vsc-frame}{HTML}{E0E0E0}

\lstdefinestyle{vscpythonlight}{
    backgroundcolor=\color{vsc-bg},
    basicstyle=\ttfamily\scriptsize\color{black},
    commentstyle=\itshape\color{vsc-comment},
    keywordstyle=\color{vsc-keyword}\bfseries,
    stringstyle=\color{vsc-string},
    numberstyle=\tiny\color{gray},
    frame=lines,
    rulecolor=\color{black},
    framerule=1pt,
    showstringspaces=false,
    tabsize=4,
    numbersep=6pt,
    breaklines=true,
    language=Python,
    alsoletter={0123456789_}, 
    emph={self,True,False,None}, emphstyle=\color{vsc-builtin},
    emph={[2]torch,nn,F}, emphstyle={[2]\color{vsc-builtin}},
    morekeywords={class,def,return,for,if,elif,else,import,from,with,as,while,try,except,raise,in,is},
    moredelim=[is][\color{vsc-builtin}]{@@}{@@}, 
}
\def\argkmax{\text{$k$-argmax}}
\def\kmax{\text{$k$-max}}

\definecolor{wacvblue}{rgb}{0.21,0.49,0.74}
\usepackage[pagebackref,breaklinks,colorlinks,allcolors=wacvblue]{hyperref}

\def\wacvPaperID{97} 
\def\confName{WACV}
\def\confYear{2027}

\title{Unifying Convolution and Attention via Convolutional Nearest Neighbors}

\author{Mingi Kang, Jeová Farias Sales Rocha Neto\\
Bowdoin College\\
255 Maine Street, Brunswick, ME, USA\\
{\tt\small \{mkang2, j.farias\}@bowdoin.edu}
}

\begin{document}
\maketitle
\begin{abstract}
Convolutional Neural Networks and Vision Transformers are the two dominant
architectural families in computer vision, defined by spatially local
convolution and global self-attention respectively. Despite their apparent
differences, we show that both operations are special cases of a single
$k$-nearest neighbor aggregation framework: convolution selects neighbors by
spatial proximity while attention selects by feature similarity, placing them at
two ends of a shared operational spectrum. We introduce \textit{Convolutional
Nearest Neighbors} (\convnn), a unified framework that exactly recovers standard
and depthwise convolution, self-attention, and sparse attention variants
including KVT-attention as special cases, and exposes the design space of
neighbor-selection strategies between them through configurable similarity
functions, positional encodings, and aggregation kernels. We validate \convnn\
on ImageNet-1K classification across two complementary architectures: a hybrid
branching layer in ResNet-50 that combines local and global feature learning,
improving top-1 accuracy by 3.0\% over the ResNet-50 baseline, and
\convnn-attention in ViT-Base that achieves 81.64\% top-1 accuracy, surpassing
standard multi-head self-attention by 0.7\%. Together, these results demonstrate
that \convnn\ provides a principled foundation for designing operations that
bridge convolutional and attention-based computation.
\end{abstract}
    
\section{Introduction}
\label{sec:intro}

Model architectures in computer vision have undergone a fundamental shift over the past two decades. Convolutional Neural Networks (CNNs) dominated through pioneering architectures like AlexNet \cite{DBLP:journals/cacm/KrizhevskySH17}, VGG \cite{DBLP:journals/corr/SimonyanZ14a}, and ResNet \cite{he2016deep}. More recently, Vision Transformers (ViT) \cite{DBLP:conf/iclr/DosovitskiyB0WZ21} have reshaped the landscape with attention-based models, achieving state-of-the-art performance across numerous vision tasks. Despite these successes in model architecture, convolution and attention are typically treated as fundamentally distinct operations. 

Traditionally, convolutions operate through local spatial aggregation, where each output is computed from a fixed spatial neighborhood or the convolutional kernel. In contrast, self-attention \cite{vaswani2017attention} enables global feature aggregation by computing similarity-based weighted combinations across all spatial positions. However, these two operations share a common principle: both operations aggregate features from selected neighbors. The distinction is in neighbor selection; convolution selects by spatial proximity, while attention selects by learned feature similarity. 

\begin{figure}
    \centering
    \includegraphics[width=1\linewidth]{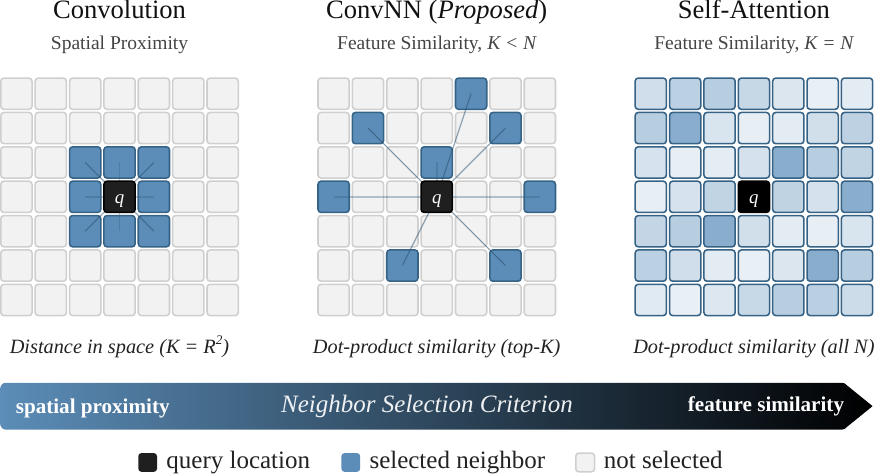}
    \caption{The \convnn operation compared to standard Convolution and Attention, showing how they live in a spectrum.}
    \label{fig:convnn_vs_att_conv}
\end{figure}

Prior work has hinted at this connection. \citet{DBLP:conf/iclr/CordonnierLJ20} demonstrated that attention with relative positional encoding act like convolution, suggesting these operations may be related. Similarly, ACmix \cite{pan2022integration} introduces a natural integration of both operations within a mixed model architecture that utilizes both convolution and self-attention through $1 \times 1$ convolutions and shift operations. However, these approaches treat convolution and attention as separate operations to be bridged externally and do not provide a solid mathematical foundation to reveal how both operations are instances of the same underlying principle. 

To formally connect convolution and attention, we introduce \textit{Convolutional Nearest Neighbors} (\convnn), a unified framework that formalizes the interconnection between convolution and attention through $k$-nearest neighbor ($k$-NN) selection. Our key insight is that both operations are specialized instances of neighbor aggregation, differing in their selection strategy. Crucially, \convnn can be formulated to recover exactly standard convolution and self-attention, proving both are points on a continuous spectrum. The flexibility of \convnn enables systematic exploration of this spectrum. Rather than choosing between convolution or attention, \convnn allows hybrid configurations that combine spatial and feature-based neighbor selection. 

We validate \convnn on ImageNet-1K classification in both architectural families. Replacing self-attention in ViT-Base, \convnn-attention improves top-1 accuracy by 0.70\% over multi-head self-attention and outperforms KVT at every tested neighbor count; in ResNet-50, a hybrid branching layer that fuses spatial and feature-similarity selection exceeds both a matched convolutional baseline and a pure-\convnn replacement. Beyond these gains, unifying the two operations exposes a design space of neighbor-selection strategies spanning convolution and attention.

Our contributions are:

\begin{itemize}
    \item A unifying k-NN formulation that casts convolution and self-attention as one neighbor-aggregation operation, differing only in how neighbors are selected — by spatial proximity or feature similarity.
    \item Exact-recovery results proving that standard and depthwise convolution (with boundary behavior under zero-padding), self-attention, KVT-attention, structured sparse attention (BigBird, Sparse Transformer, causal masking), and multi-head topologies are all special cases of this framework.
    \item Practical realizations of \convnn: random sparse candidate search, a fused Triton kernel that lowers forward peak memory by $2.3\times$, and a hybrid branching layer combining both selection modes.
    \item A controlled empirical study on ImageNet-1K isolating the effect of \emph{how} neighbors are selected at matched neighbor budgets, across both CNN and Transformer backbones.
 \end{itemize}

\section{Related Works}
\label{sec:related}

\paragraph{Connections between Convolution and Attention.}
Recent works began exploring the theoretical relationship between these seemingly distinct operations. \citet{DBLP:conf/iclr/CordonnierLJ20} provides evidence that self-attention with relative positional encoding can express any convolutional operation, highlighting that convolution and attention may exist on a spectrum. \citet{pan2022integration} reformulate convolution and self-attention through $1 \times 1$ convolutions and shifted feature maps, offering a unified treatment that leverages both operations. \textit{Non-Local Neural Networks} \cite{DBLP:conf/cvpr/0004GGH18} and \textit{SENet} \cite{DBLP:journals/pami/HuSASW20} introduced capturing long-ranged dependencies in convolutional blocks, demonstrating that operations that combine both local and global feature learning enhance performance in deep neural networks. \textit{Attention Augmented Convolutional Networks} \cite{DBLP:conf/iccv/BelloZLVS19} augment convolutional layers with relative self-attention layers for visual recognition tasks. \textit{Convolution-enhanced image Transformer (CeiT)} \cite{DBLP:conf/iccv/YuanG0ZYW21} enhances Vision Transformers with architectural modifications to incorporate locality with convolutions. \textit{CoAtNet} \cite{DBLP:conf/nips/DaiLLT21} systematically analyzes depthwise convolutions and self-attention, demonstrating that both operate as per-dimension weighted sums within a predefined receptive field. \textit{ConViT} \cite{DBLP:conf/icml/dAscoliTLMBS21} introduces gated positional self-attention, which utilizes a gating parameter to dynamically balance the expressive properties of convolutions and self-attention. \textit{CvT} enhances Vision Transformers through the introduction of Convolutional Token Embeddings and Convolutional Projections, bridging the gap between Transformers and CNNs. Notably, \citet{katharopoulos2020transformers} observe that the Linear Attention formulation is equivalent to a recurrent neural network in the sequential setting. \citet{han2021connection} further showed that local attention in Vision Transformers is structurally equivalent to depthwise convolution under shared properties of sparse local connectivity and per-channel weight sharing. These results establish important pairwise links (attention can express convolution \cite{DBLP:conf/iclr/CordonnierLJ20}, local attention coincides with depthwise convolution \cite{han2021connection}, and hybrid designs interleave the two \cite{pan2022integration, DBLP:conf/nips/DaiLLT21}) but each connects a specific pair of operations in a single direction. None provides one operation whose configuration recovers both convolution and a swath of attention variants as instances of a common selection principle.

\vspace{-10pt}
\paragraph{$k$-Nearest Neighbors in Deep Learning}
The $k$-Nearest Neighbor ($k$-NN) principle has recently reemerged as a valuable design component in neural architectures. \textit{Neural Nearest Neighbors Networks} \cite{DBLP:conf/nips/Plotz018} first explored the integration of $k$-NN reasoning primarily for few-shot learning. More recently, $k$-NN has been incorporated into Transformers to mitigate the quadratic computational and memory cost of self-attention. In particular, \textit{$k$-NN Attention for Boosting Vision Transformers (KVT)} \cite{DBLP:conf/eccv/WangWWLCLJ22} enhances Vision Transformers by introducing a $k$-NN attention mechanism that restricts each token's attention to its most relevant neighbors rather than the entire sequence. Similarly, \textit{Unlimiformer} \cite{DBLP:conf/nips/Bertsch0NG23} addresses cross-attention computational complexity by using $k$-NN  index, only computing the scores to the top-$k$ tokens. \textit{Routing Transformers} \cite{roy-etal-2021-efficient} rely on $k$-means clustering to model sparse attention matrices. \textit{Memory-efficient Transformers} rely on top-$k$ scores within chunks of queries, resulting in memory savings for large sequences. In all of these, k-NN is an efficiency mechanism layered onto attention to prune the candidate set. We instead treat neighbor selection as the defining axis of the operation itself: the criterion (spatial versus feature similarity) and breadth ($k$) are the parameters that distinguish convolution from attention, making 
$k$-NN the abstraction that unifies them rather than an optimization applied to one of them.

\section{Background}
\label{sec:background}

\begin{figure*}[t]
    \centering
    \includegraphics[width=1.0\linewidth]{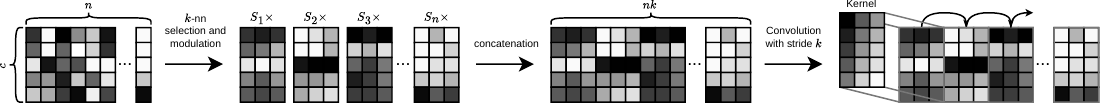}
    \caption{The \convnn operation}
    \label{fig:convnn_diagram}
\end{figure*}

\subsection{Notation}
\label{subsec:notation}
Scalar variables, vectors, and matrices are denoted in plain lowercase (e.g., $x$), bold lowercase (e.g., $\mathbf{x}$), and plain uppercase (e.g., $X$), respectively. High-order tensors are denoted by bold uppercase letters (e.g., $\mathbf{X}$). The shape of a tensor is written as $\mathbf{X} \in \mathbb{R}^{a \times b \times c}$, abbreviated as $[a, b, c]$ where convenient. A vector of ones in $\mathbb{R}^k$ is denoted as $\mathbf{1}_k$. 

We adopt Python-style indexing, where square brackets indicate row and column selection. Given a matrix $A \in \mathbb{R}^{N \times M}$, $A[i, :]$ selects the $i$-th row of $A$. We use $[n]$, for $n$ natural, to denote the set $\{0,\ldots, n-1\}$. Given an index sets $I \subset [N]$, $A[I, :]$ selects the rows of $A$ corresponding the indices in $I$. 

A vector $\mathbf{x}$ is $\ell_2$-normalized when $\|\mathbf{x}\|_2 = 1$. The operator $\operatorname{softmax}(\mathbf{x})$ normalizes $\mathbf{x}$ into a probability distribution with $\operatorname{softmax}(\mathbf{x})_i = \exp(x_i) / \sum_j \exp(x_j)$. The operator $\operatorname{diag}(\mathbf{v})$ denotes the diagonal matrix with entries of $\mathbf{v}$ on its main diagonal. 

\subsection{$k$-Nearest Neighbors}
\label{subsec:knn_notation}
Let $X \in \mathbb{R}^{N \times d}$ be a matrix whose $i$-th row $\mathbf{x}_i \in \mathbb{R}^d$ is the $i$-th feature vector, for $i = 1, \dots, N$. We make use of two operators throughout: $\kmax$ and $\argkmax$. $\kmax : \mathbb{R}^{N \times N} \to \mathbb{R}^{N \times K}.$ returns the $K$ largest values in each row of a matrix and  $\argkmax : \mathbb{R}^{N \times N} \to [N]^{N \times K}$ returns the column indices of these $K$ largest values for each row.

Given a similarity matrix $S \in \mathbb{R}^{N \times N}$, the global index matrix $I$ and top-$K$ similarity matrix $T$ are defined as:
\begin{equation*}
    I = \argkmax(S) \in [N]^{N \times K}, \quad   T = \kmax(S) \in \mathbb{R}^{N \times K},
\end{equation*}
where $I_i = I[i, :]$ and $T_i = T[i, :]$ denote the neighbor indices and similarity values for query $\mathbf{x}_i$, respectively. The set of $K$-nearest neighbors of $\mathbf{x}_i$ is then retrieved as the submatrix $X[I_i, :] \in \mathbb{R}^{K \times d}$.

\subsection{Convolution}
\label{subsec:convolution}
A convolutional layer applies a learned kernel that sweeps across the input spatially, computing a weighted sum of local input values at each position to produce a single output. Depthwise convolution applies an independent kernel per channel with no cross-channel mixing. 1D convolution (\textsc{Conv1D}) operates over a single spatial axis on inputs of shape $[N \times C]$, commonly applied to sequential data such as audio and token sequences. 2D convolution (\textsc{Conv2D}) extends this to two spatial axes on inputs of shape $[H \times W \times C]$ with a $R \times R$ kernel, and is the standard operation for 2D image feature extraction. 

\subsection{Self-Attention}
\label{subsec:self_attention}
Self-attention~\cite{vaswani2017attention} lets each position of an input
sequence $X \in \mathbb{R}^{N \times d}$ attend to all $N$ positions. The input
is mapped to queries, keys, and values through learned projections
$Q = XW^Q$, $K = XW^K$, and $V = XW^V$, with
$W^Q, W^K \in \mathbb{R}^{d \times d_k}$ and $W^V \in \mathbb{R}^{d \times d_v}$.
The output is given by scaled dot-product attention:
\begin{equation}
    \operatorname{Attention}(Q, K, V)
    = \operatorname{softmax}\!\left(\frac{QK^\top}{\sqrt{d_k}}\right)V,
    \label{eq:self_attention}
\end{equation}
\noindent where the softmax is applied row-wise and $d_k$ is the dimension of
the queries and keys; the $1/\sqrt{d_k}$ factor counteracts large dot products
that would otherwise push the softmax into low-gradient regions. Unlike
convolution, attention imposes no locality constraint and can model global
dependencies directly, at $O(N^2 d)$ time and $O(N^2)$ memory cost.

\section{Convolutional Nearest Neighbors Framework}
\label{sec:core_framework}

We now present \convnn, an operational framework that generalizes both convolution and self-attention, and is built upon three core steps: \textit{similarity computation}, \textit{neighbor selection and modulation}, and \textit{weighted aggregation}. 

Let $X \in \mathbb{R}^{N \times C}$ denote the input feature matrix, where $N$ is the number of feature vectors and $C$ is the channel dimension. For inputs with spatial structure of shape $[H, W, C]$, \convnn\ first flattens the spatial dimensions to obtain a matrix of shape $[HW, C]$, applies the procedure described below, and reshapes the result back to $[H, W, C']$. The input $X$ may augment raw features with spatial information by appending normalized positional coordinates along the channel dimension, enabling \convnn\ to blend feature-based and spatially-aware similarity, a property that is central to its connection with convolution, as detailed in Section~\ref{sec:convnn_conv_connection} and the supplementary material. An overview of the \convnn\ operation is illustrated in Figure~\ref{fig:convnn_diagram} and its PyTorch implementation is depicted in Algorithm \ref{code:convnn_basic}.

\begin{lstlisting}[caption={PyTorch implementation of the \convnn operation},
    basicstyle=\scriptsize\ttfamily,
    captionpos=t,
    label={code:convnn_basic}
],
class ConvNN(nn.Module):
    def __init__(self, C, C_qk, C_v, C_out, K):
        """
        Args:
            C     : input channel dimension
            C_qk  : projected query/key dimension
            C_v   : projected value dimension
            C_out : output channel dimension (C')
            K     : number of nearest neighbors
        """
        super(ConvNN, self).__init__()
        self.K    = K
        self.f_Q  = nn.Linear(C, C_qk)
        self.f_K  = nn.Linear(C, C_qk)
        self.f_V  = nn.Linear(C, C_v)
        self.rho = nn.Softmax(dim=-1)
        self.conv = nn.Conv1d(
            in_channels=C_v,
            out_channels=C_out,
            kernel_size=K,
            stride=K
        )

    def similarity(self, Q, K):
        return Q @ K.transpose(-2, -1)
        
    def forward(self, X):
        B, N, C = X.shape
        Q, K, V = self.f_Q(X), self.f_K(X), self.f_V(X)

        ## Step 1 - Similarity Computation
        S = self.similarity(Q, K)               

        ## Step 2 - Neighbor Selection and Modulation
        # s_i = kmax_k(S)[i, :],  
        # I_i = argkmax_k(S)[i, :]
        s, I = torch.topk(S, k=self.K, dim=-1)  
        s = self.rho(s) 
        
        # Gather V[I_i, :] in R^{K x C_v}
        C_v   = V.shape[-1]
        V_t   = V.permute(0, 2, 1)                          
        V_exp = V_t.unsqueeze(-1).expand(B,C_v,N,self.K)  
        I_exp = I.unsqueeze(1).expand(B, C_v, N, self.K)    
        V_nn  = torch.gather(V_exp, dim=2, index=I_exp)   
        
        # Perform X_nn_i = diag(rho(s_i)) * V[I_i, :]
        s_exp = s.unsqueeze(1).expand(B, C_v, N, self.K)     
        X_nn  = s_exp * V_nn             
        
        ## Step 3 - Weighted Aggregation
        # Reshape to [B, C_v, N*K] for Conv1D with kernel_size=K, stride=K
        X_nn = X_nn.permute(0,1,3,2).reshape(B,C_v,-1)  
        Y    = self.conv(X_nn).permute(0, 2, 1)                              
        return Y
\end{lstlisting}

\paragraph{1. Similarity Computation.}
Let $f_Q : \mathbb{R}^{N \times C} \to \mathbb{R}^{N \times C_{qk}}$ and
$f_K : \mathbb{R}^{N \times C} \to \mathbb{R}^{N \times C_{qk}}$ be projection
functions, and define $Q = f_Q(X)$ and $K = f_K(X)$, where $C_{qk}$ is the
shared dimension of the query and key spaces. A similarity function
$\operatorname{sim}(q_i, k_j)$ scores how closely key $k_j$ matches query
$q_i$, computed in the projected space $\mathbb{R}^{C_{qk}}$ rather than the
original feature space $\mathbb{R}^C$. This yields a similarity matrix
$S \in \mathbb{R}^{N \times N}$,
\begin{equation}
    S_{ij} = \operatorname{sim}(q_i, k_j),
\end{equation}
\noindent whose scores drive neighbor selection in the next step. By default
$\operatorname{sim}$ is the dot product, in which case $S = QK^\top$. Other
comparison functions may be substituted: cosine similarity applies directly,
while a distance metric $D(q_i, k_j)$ is first converted to a similarity, e.g.\
$\operatorname{sim}(q_i, k_j) = \exp(-D(q_i, k_j)^2)$, so that more similar pairs
receive larger scores.


\paragraph{2. Neighbor Selection and Modulation.}
Let $f_V : \mathbb{R}^{N \times C} \to \mathbb{R}^{N \times C_v}$ be a projection function, and define $V = f_V(X)$. Using the operators defined in Section~\ref{subsec:knn_notation}, we compute the top-$K$ similarity values and their corresponding indices for each query: 
\begin{equation*}
    \mathbf{s}_i = T[i, :] = \kmax(S)[i, :] \in \mathbb{R}^{K}, 
\end{equation*}
\begin{equation*}
    I_i = I[i, :] = \argkmax(S)[i, :] \in [N]^{K}.
\end{equation*}
A weighting function $\rho : \mathbb{R}^K \to \mathbb{R}^K$ maps these raw scores
to aggregation weights, instantiated as uniform weighting
($\rho(\mathbf{z}) = \mathbf{1}_K$) or softmax normalization
($\rho(\mathbf{z}) = \operatorname{softmax}(\mathbf{z})$); both sum to one, so the
two choices remain on a common scale. Letting
$\Lambda_i = \operatorname{diag}(\rho(\mathbf{s}_i))$, the \textit{neighborhood
matrix} for $\mathbf{x}_i$ is
\begin{equation}
    X_{\text{nn}, i} = \Lambda_i\, V[I_i, :] \in \mathbb{R}^{K \times C_v},
    \label{eq:neigh_mat}
\end{equation}
\noindent collecting the $K$ projected neighbors of $\mathbf{x}_i$, each scaled
by its weight in $\rho(\mathbf{s}_i)$.

\paragraph{3. Weighted Aggregation.}
Let $X_{\text{nn}} \in \mathbb{R}^{(KN) \times C_v}$ denote the concatenation of all neighborhood matrices $\{X_{\text{nn},i}\}_{i=1}^N$ along the row dimension. \convnn\ applies a \textsc{Conv1D} operation (standard or depthwise) with $C'$ output channels, kernel size $K$, and stride $K$ to $X_{\text{nn}}$, yielding:

\begin{equation}
    Y = \text{Conv1D}(X_{\text{nn}}) \in \mathbb{R}^{N \times C'}.
\end{equation}

\section{Practical Implementations}

\subsection{Sparse Candidate Search} 
\label{sec:sparse_candidate}
Exhaustive $k$-NN search over all $N$ input features incurs $O(N^2)$ similarity computations, which is prohibitive for large feature maps. To reduce this cost, we restrict the search to a random candidate set of size $r \ll N$, reducing the similarity computation and memory footprint from $O(N^2)$ to $O(Nr)$.

Formally, let $I_r \subset [N]$ be a set of $r$ indices drawn uniformly at random without replacement. The sparse candidate set is: 
\begin{equation}
    \mathbf{X}^{\text{sparse}} = \mathbf{X}[I_r, :] \in \mathbb{R}^{r \times C}.
\end{equation}
We deliberately choose random selection over structured alternatives for two reasons. First, it is simple, requires no spatial assumptions, and applies identically to 1D and 2D inputs with no additional hyperparameters. Second, and more importantly, structured spatial subsampling selects candidates at regular spatial intervals, which reintroduces a spatial proximity bias into neighbor selection. The most similar candidates are often the spatially nearest ones, causing \convnn\ to degrade toward convolution-like behavior. Random selection avoids this by sampling candidates uniformly across the full spatial extent, a strategy supported both theoretically~\cite{zaheer2020big} and empirically in vision transformers~\cite{zhang2023vision}, ensuring that the neighbor pool is not dominated by locally adjacent features.

\subsection{Fast Convolutional Nearest Neighbors with Triton Acceleration}
\label{sec:triton_convnn}
The standard \convnn\ formulation introduced in Section~\ref{sec:core_framework} incurs substantial memory overhead at training and inference time, as its three-step pipeline (gathering neighbors from $V$, applying attention weights $\rho(\mathbf{s}_i)$, and aggregating via a \textsc{Conv1D} kernel) requires materializing several large intermediate tensors in GPU memory, creating significant pressure during the backward pass. We address this by introducing a fused Triton kernel~\cite{10.1145/3315508.3329973} that merges all three steps into a single GPU kernel launch, eliminating the intermediate tensors and allowing the backward pass to recompute required values on the fly rather than storing them. Table~\ref{tab:fastconvnn_speed} reports forward and backward pass times and peak memory usage for a representative configuration. The detailed design and implementation are available in the supplementary material. 

\begin{table}[H]
    \centering
    \footnotesize
    \vspace{5pt}
    \caption{Speed and memory benchmark of standard PyTorch versus Triton-fused \convnn\ aggregation on a single H100 GPU. Configuration: batch size 32, sequence length 197, embedding dimension 768, 12 attention heads, $K = 8$, depthwise convolution.}
    \begin{tabularx}{\linewidth}{Xccc}
        \toprule
        \textbf{Metric} & \textbf{Original} & \textbf{Triton} & \textbf{Improvement} \\
        \midrule
        Forward time (ms)         & 4.92   & 4.24   & $1.16\times$ faster \\
        Backward time (ms)        & 9.29   & 6.84   & $1.36\times$ faster \\
        Forward peak memory (MB)  & 795.01 & 341.95 & $2.32\times$ lower  \\
        Backward peak memory (MB) & 904.88 & 657.54 & $1.38\times$ lower  \\
        \bottomrule
    \end{tabularx}
    \label{tab:fastconvnn_speed}
\end{table}

\subsection{Hybrid Branching Layer}
\label{subsec:hybrid_branching_layer}
A key distinction between \convnn\ and standard convolution lies in their receptive behavior: convolution aggregates information from a fixed local spatial window, whereas \convnn\ selects neighbors based on feature similarity rather than spatial proximity. Building on prior work that integrates local and global feature modeling~\cite{szegedy2015going, peng2021conformer, chen2022regionvit}, we introduce a hybrid \textit{branching layer} that processes the input through two parallel branches, convolution and \convnn, with output channels allocated according to a scalar parameters $\lambda \in [0, 1]$. The two branch outputs are concatenated along the channel dimension and mixed via a pointwise ($1\times 1$) convolution. This design is analogous to the multi-scale parallel branch structure of the Inception module~\cite{szegedy2015going}, but replaces the multi-kernel convolution branches with a local-global pair. Random sparse candidate sampling is applied within the \convnn\ branch, encouraging it to search non-locally across the image rather than relying solely on spatially or semantically adjacent neighbors. 

\begin{figure}
    \centering
    \includegraphics[width=1.0\linewidth]{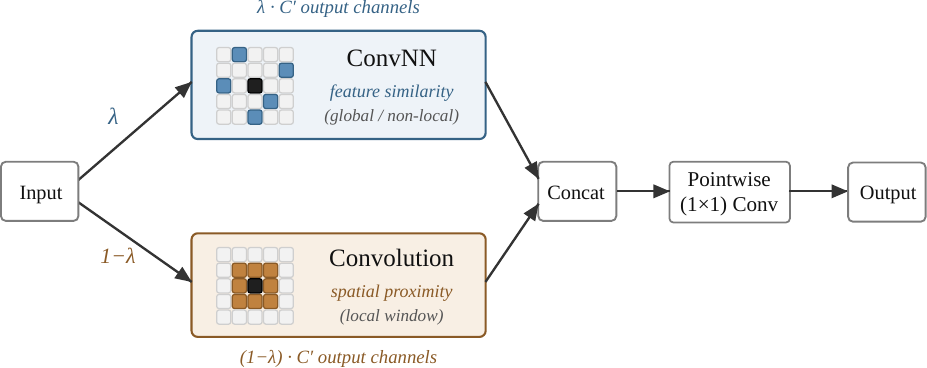}
    \caption{Hybrid Branching Layer combining \convnn\ and standard convolution with branch ratio $\lambda$.}
    \label{fig:hybrid_branching_layer}
\end{figure}

\section{Theoretical Analysis}
\label{sec:theory}
We now establish the theoretical connection of \convnn\ to convolution and
attention, demonstrating how \convnn\ generalizes both operations. We first show
that standard and depthwise convolution arise as special cases under
spatial-coordinate similarity (\Cref{sec:convnn_conv_connection}), then that
self-attention and its sparse and multi-head variants arise under scaled
dot-product similarity (\Cref{sec:convnn_attn_connection}).

\subsection{Connection to Convolution}
\label{sec:convnn_conv_connection}
By restricting the query and key projections to raw spatial coordinates and
measuring similarity by spatial distance, neighbor selection in \convnn\ reduces
to a fixed local window, recovering convolution exactly.

\begin{theorem}[Equivalence to Convolutional Operations]
\label{thm:conv_equivalence}
Let $\mathbf{X}$ be an input tensor augmented with normalized spatial
coordinates. If the query and key projections are restricted to the raw spatial
coordinates, the similarity metric is the exponential of the negative squared Euclidean distance
$S_{ij} = \exp(-\|p_i - p_j\|_2^2)$, and neighbor selection retrieves $K = R^d$
neighbors (with $R$ an odd kernel size and $d$ the spatial dimension), then
\convnn\ under uniform weighting $\rho(\mathbf{z}) = \mathbf{1}_K$ exactly
recovers standard and depthwise convolution at all interior spatial positions.
\end{theorem}

\begin{theorem}[Exact Boundary Recovery via Zero-Padding]
\label{thm:padding_equivalence}
The equivalence of \Cref{thm:conv_equivalence} holds universally across all
spatial positions, including sequence boundaries, under a zero-padding scheme
matching the standard convolutional receptive field.
\end{theorem}

\begin{proof}[Proof of \Cref{thm:conv_equivalence,thm:padding_equivalence}]
Deferred to the supplementary material, where we give the complete constructive
proofs for both 1D and 2D inputs, together with a characterization of the
boundary-skewing effects that arise in unpadded sequences.
\end{proof}

\subsection{Connection to Attention}
\label{sec:convnn_attn_connection}
Replacing the spatial-coordinate similarity with a scaled dot product over
learned projections, the same neighbor-aggregation operation recovers dense
global attention, its sparse variants, and the multi-head topology.

\begin{theorem}[Generalization of Attention Frameworks]
\label{thm:attention_generalization}
Let $Q, K, V$ be projections of an input $X$ via learned linear maps, and let the
similarity matrix $S$ be the scaled dot product. Then:
\begin{enumerate}
    \item \textbf{Standard self-attention} is exactly recovered when neighbor
    selection is global ($K = N$) and the depthwise aggregation weights are
    unit-valued ($w_{k, m} = 1$).
    \item \textbf{KVT-Attention}~\cite{DBLP:conf/eccv/WangWWLCLJ22} is exactly
    recovered when $K < N$ with unit aggregation weights.
    \item \textbf{Structured sparse attention} (e.g., BigBird, Sparse
    Transformer) is exactly recovered by injecting a mask
    $M \in \{0, -\infty\}^{N \times N}$ into the similarity, $S' = S + M$.
\end{enumerate}
\end{theorem}

\begin{theorem}[Equivalence under Multi-Head Topology]
\label{thm:multihead_equivalence}
Folding the head dimension into the batch dimension prior to neighbor selection,
a single \convnn\ operation over $\mathbb{R}^{BH \times N \times d_k/H}$ is
algebraically equivalent to running $H$ independent attention heads in parallel.
\end{theorem}

\begin{proof}[Proof of \Cref{thm:attention_generalization,thm:multihead_equivalence}]
Deferred to the supplementary material, which provides the
index-level expansion, the explicit sparse-mask constructions (including the
causal autoregressive pattern), and the multi-head folding argument.
\end{proof}
\section{Experiments}
In this section, we validate \convnn\ across the spectrum of Convolutional Neural Networks and Vision Transformers on ImageNet-1K image classification. We standardize our training configurations and benchmark \convnn\ against existing architectures that generalize convolution and attention. Additionally, our summary tables report the performance metrics of several established baselines as cited in the literature.

\subsection{Experimental Setup}

We evaluate on ImageNet-1K~\cite{ILSVRC15} (1.28M training, 50K validation, 1,000 classes), following the Swin Transformer training protocol~\cite{DBLP:conf/iccv/LiuL00W0LG21}. Training augmentations include random resized cropping, horizontal flipping, RandAugment ($m=9$, $n=2$)~\cite{cubuk2020randaugment}, color jittering, random erasing, MixUp ($\alpha = 0.8$)~\cite{zhang2017mixup}, and CutMix ($\alpha = 1.0$)~\cite{yun2019cutmix}. Validation images are resized with bicubic interpolation, center-cropped to $224 \times 224$, and normalized using ImageNet statistics. 

All models are trained for 300 epochs with AdamW~\cite{loshchilov2017decoupled} (lr$=10^{-3}$, wd$=0.05$, $\beta=(0.9, 0.999)$), cosine annealing with 20-epoch linear warmup~\cite{loshchilov2016sgdr} (from $10^{-7}$, min lr$=5 \times 10^{-6}$), and soft-target cross-entropy loss~\cite{hugger2024towards} with label smoothing $\epsilon = 0.1$. Experiments run on four NVIDIA H100 GPUs on Indiana Jetstream2~\cite{10.1145/3437359.3465565} with Distributed Data Parallel, global batch size 1,280, bfloat16 AMP, and TF32 enabled.

\subsection{Results on Convolutional Neural Networks}

\subsubsection{Model Setup}
For convolutional experiments, we use a ResNet-50 backbone \cite{he2016deep} with hybrid branching layer ($\lambda = 0.5$) replacing all standard $3 \times 3$ convolutions with stride 1. The \convnn branch uses identity projections for $Q, K, V$, cosine similarity, uniform weights ($\rho(\mathbf{z}) = \mathbf{1}_K$), and standard \textsc{Conv1D} for weighted aggregation. The standalone \convnn\ baseline is direct per-layer replacements without the branching module and therefore exclude the pointwise $1 \times 1$ mixing convolution. 

\subsubsection{Effect of Neighbor Count and Candidate Selection in ResNet-50}

\begin{figure}[t]
\centering
\begin{tikzpicture}
\begin{axis}[
    width=1.0\linewidth, height=6cm,
    xlabel={Neighbor count $k$},
    ylabel={Top-1 Accuracy (\%)},
    xtick={4,9,16},
    xticklabels={$4$,
                 $9$,
                 $16$},
    xticklabel style={font=\small, align=center},
    ymin=75.8, ymax=80.5,
    ymajorgrids=true, grid style={dotted, gray!40},
    legend style={at={(0.98,0.22)}, anchor=south east, font=\scriptsize,
                  draw=gray!40, fill=white, row sep=2pt},
    tick label style={font=\small},
    label style={font=\small},
]
\addplot[color=blue!70, mark=*, thick, mark size=2.5pt]
    coordinates {(4,79.06) (9,79.42) (16,79.76)};
\addlegendentry{Branch (All candidates)}

\addplot[color=teal!80, mark=triangle*, thick, dashed, mark size=2.5pt]
    coordinates {(4,78.73) (9,78.98) (16,79.12)};
\addlegendentry{Branch (Rand, $r{=}32$)}

\addplot[color=red!70, thick, densely dashed,
         mark=*, mark size=2pt,
         mark repeat=1]
    coordinates {(4,76.22) (9,76.22) (16,76.22)};
\addlegendentry{\convnn\ only ($k{=}9$)}

\addplot[color=gray!60, thick, dotted,
         mark=square*, mark size=1.5pt,
         mark repeat=1]
    coordinates {(4,76.7) (9,76.7) (16,76.7)};
\addlegendentry{ResNet-50 baseline}

\end{axis}
\end{tikzpicture}
\caption{Top-1 accuracy on ImageNet-1K for ResNet-50 with hybrid branching ($\lambda=0.5$) across neighbor counts $k$, comparing all-feature and random sparse ($r=32$) candidate selection. The \convnn-only and ResNet-50 entries are standalone layer replacements without the branching module.}
\label{fig:imagenet_resnet_k}
\end{figure}

Figure~\ref{fig:imagenet_resnet_k} compares the hybrid branching layer and pure \convnn\ replacement against the ResNet-50 baseline across neighbor counts $k$, isolating the effect of how neighbors are selected at matched architectural depth. 

Three observations emerge. First, the hybrid branching layer with all-feature selection consistently outperforms the ResNet-50 baseline (76.7\%) at every value of $k$, with accuracy increasing monotonically from 79.06\% at $k=4$ to 79.76\% at $k=16$, a gain of 3.06\% at peak. This improvement is achieved without adding a new architectural stage, since the \convnn branch replaces rather than augments the convolutional branch. Second, the pure \convnn\ replacement at $k=9$ achieves 76.22\%, matching around the ResNet-50 baseline but providing no gain, confirming that feature-similarity selection alone is not sufficient and that the combination of local and global aggregation in the branching layer is what drives the improvement. Third, sparse random selection with $r=32$ underperforms all-feature selection across all $k$ values but consistently outperforms the ResNet-50 baseline, achieving 78.73\% at $k=4$ and 79.12\% at $k=16$, making it a competitive operating point when computational cost is constrained. 

\begin{table}[t]
\vspace{5pt}
\centering
\caption{ImageNet-1K top-1 classification accuracy for convolutional architectures. Models marked with $*$ utilize a ResNet-50 backbone where standard $3 \times 3$ convolutions are replaced by the specified module. For our \convnn\ configurations, (A) denotes all-feature candidate selection, and (R) denote random sparse candidate selection with $r=32$. ``Branch`` indicates the hybrid convolution and \convnn\ branching layer. Baseline results for ResNet, EfficientNet, and RegNet architectures are reported as evaluated in \citet{DBLP:conf/iccv/YuanG0ZYW21}. ConvNeXt performance metrics are taken from the original publication \citet{liu2022convnet}.}
\footnotesize
\setlength{\tabcolsep}{4pt}
\begin{tabular}{lcccc}
\toprule 
Model & \#Param. & Image Size & FLOPs (G) & Top-1 \\
\midrule 
$\ast$ \convnn (A, $k=9$)                       & 26M  & $224^2$ & 10.3  & 76.2 \\
$\ast$ Branch (A, $k=9$)                        & 26M  & $224^2$ & 10.7 & 79.4 \\
$\ast$ Branch (A, $k=16$)                       & 29M  & $224^2$ & 10.7 & 79.8 \\
$\ast$ Branch (R, $k=9$)                        & 26M  & $224^2$ & 5.6  & 79.0 \\ 
$\ast$ Branch (R, $k=16$)                       & 29M  & $224^2$ & 5.6  & 79.1 \\
\midrule 
ResNet-50 \cite{he2016deep}                     & 26M  & $224^2$ & 4.1  & 76.7 \\
ResNet-101 \cite{he2016deep}                    & 45M  & $224^2$ & 7.8  & 78.3 \\
ResNet-152 \cite{he2016deep}                    & 60M  & $224^2$ & 11.5 & 78.9 \\
\midrule
RegNetY-4GF \cite{radosavovic2020designing}     & 21M  & $224^2$ & 4.0  & 80.0 \\
RegNetY-8GF \cite{radosavovic2020designing}     & 39M  & $224^2$ & 8.0  & 81.7 \\
RegNetY-16GF \cite{radosavovic2020designing}    & 84M  & $224^2$ & 16.0 & 82.9 \\
EfficientNet-B0 \cite{tan2019efficientnet}      & 5M   & $224^2$ & 0.4  & 77.1 \\
\midrule 
ConvNeXt-T \cite{liu2022convnet}                & 29M  & $224^2$ & 4.5  & 82.1 \\
ConvNeXt-S \cite{liu2022convnet}                & 50M  & $224^2$ & 8.7  & 83.1 \\
ConvNeXt-B \cite{liu2022convnet}                & 89M  & $224^2$ & 15.4 & 83.8 \\
ConvNeXt-L \cite{liu2022convnet}                & 198M & $224^2$ & 34.4 & 84.3 \\
\bottomrule
\end{tabular}
\label{tab:imagenet_resnet_results}
\vspace{10pt}
\end{table}

\subsection{Results on Vision Transformers}
\subsubsection{Model Setup}
For attention experiments, we replace the standard multi-head self-attention in ViT-Base backbone with \convnn, using fully learned $Q, K, V$ linear projections, scaled dot-product similarity, $\rho(\mathbf{z}) = \operatorname{softmax}(\mathbf{z})$, and depthwise Conv1D with initialized weights to uniform of 1. In the ViT experiments, we include the Fast-\convnn Triton-accelerated kernel variants for larger $k$. The experiments use ViT-Base configuration with patch size of 16, 12 layers, 12 attention heads, 768 hidden dimension, and 3072 MLP dimension.

\subsubsection{Learned vs.\ Fixed Aggregation Across Neighbor Counts}

\begin{figure}[t]
\centering
\begin{tikzpicture}
\begin{axis}[
    width=0.85\linewidth, height=5.5cm,
    xlabel={Neighbor count $k$},
    ylabel={Top-1 Accuracy (\%)},
    xtick={9,16,25,36,100},
    xticklabels={9,16,25,36,100},
    ymin=80.3, ymax=82.0,
    ymajorgrids=true, grid style={dotted, gray!40},
    legend style={at={(0.98,0.02)}, anchor=south east, font=\scriptsize,
                  draw=gray!40, fill=white, row sep=1pt},
    tick label style={font=\small},
    label style={font=\small},
]
\addplot[color=blue!70, mark=*, thick, mark size=2pt]
    coordinates {(9,81.19) (16,81.60) (25,81.64) (36,81.61)};
\addlegendentry{\convnn (All)}

\addplot[color=teal!80, mark=triangle*, thick, dashed, mark size=2.5pt]
    coordinates {(9,80.50) (16,81.04) (25,81.28) (36,81.03) (100,81.04)};
\addlegendentry{KVT}

\addplot[color=gray!90, thick, dotted, mark=none]
    coordinates {(9,80.94) (16,80.94) (25,80.94) (36,80.94) (100,80.94)};
\addlegendentry{MHSA baseline}
\end{axis}
\end{tikzpicture}
\vspace{-5pt}
\caption{Top-1 accuracy on ImageNet-1K for ViT-Base comparing
         \convnn and KVT (both
         NH=12) across neighbor counts $k$. 
         Results at $k = 25$ and $k = 36$ use the Fast-\convnn\ implementation with a custom Triton kernel from Section~\ref{sec:triton_convnn}.}
\label{fig:imagenet_vit_k}
\end{figure}

Figure~\ref{fig:imagenet_vit_k} compares \convnn-attention and KVT-attention (both with 12 heads) across varying $k$ under our standardized training protocol. The key architectural difference between them is the aggregation step: KVT fixes the depthwise \textsc{Conv1D} weights to unit values, while \convnn\ initializes them to 1 but allows them to be learned, enabling adaptive per-neighbor weighting. 

\convnn\ consistently outperforms KVT at every matched $k$ value, with the gap widening at larger neighborhoods at 81.19\% vs. 80.50\% at $k=9$, and 81.64\% vs. 81.28\% at $k=25$. The Fast-\convnn\ Triton variant pushes top-1 accuracy to 81.64\% 
at $k=25$, a gain of 0.70\% over our MHSA baseline trained under identical conditions (80.94\%), and 3.74\% over the originally reported ViT-B/16 result of 77.9\% trained at $384^2$ resolution with large-scale pretraining~\cite{DBLP:conf/iclr/DosovitskiyB0WZ21}. KVT accuracy peaks at $k=25$ (81.28\%) and degrades for larger $k$, while \convnn\ remains stable from $k=25$ to $k=36$ (81.64\% to 81.61\%). This divergence suggests that the learned aggregation kernel is more effective at exploiting large neighborhoods than the fixed uniform-weight sum used in KVT; as $k$ grows, the ability to assign non-uniform importance to each of the $K$ selected neighbors becomes increasingly valuable, and KVT's inability to learn this weighting leads to its degration at larger $k$.

\subsubsection{Training Dynamics and Late-Stage Convergence}

\begin{table}[t]
\centering
\scriptsize
\vspace{4pt}
\setlength{\tabcolsep}{4pt}
\caption{Top-1 accuracy on ImageNet-1K at 50-epoch intervals for ViT-Base with NH=12. \convnn use all-feature candidate selection.}
\label{tab:imagenet_vit_convergence}
\begin{tabular}{lcccccc}
\toprule
Layer Type & Ep. 50 & Ep. 100 & Ep. 150 & Ep.200 & Ep. 250 & Ep. 300 \\
\midrule
MHSA                   & \textbf{71.70} & \textbf{76.72} & \textbf{78.69} &  79.96 & 80.95 &  80.94 \\
\midrule[0.2pt]
\convnn ($k=9$)        & 67.57 & 73.61 &  76.38 & 78.63 & 80.74 & 81.19    \\   
\convnn ($k=25$)       & 69.83 & 74.91 & 77.79 &  \textbf{80.13} & \textbf{81.26} & \textbf{81.64} \\
\midrule[0.2pt]
KVT ($k=9$)            & 65.25 & 71.31 & 75.04 & 78.13 & 79.84 & 80.50 \\    
KVT ($k=25$)           & 69.79 & 74.75 & 77.71 & 79.80 & 80.81 & 81.28 \\
\bottomrule
\end{tabular}
\vspace{5pt}
\end{table}

Table~\ref{tab:imagenet_vit_convergence} reports top-1 accuracy at 50-epoch intervals across 300 epochs of training. A clear pattern emerges for \convnn, it lags behind the MHSA baseline in the early and middle stages of training, but closes the gap and surpasses it in the final epochs. This behavior is explained by the role of the depthwise Conv1D aggregation kernel. The kernel weights are originally initialized to 1 and is learned during training. When the learning rate is high with large gradient updates in the early training stages, the kernel weights undergo exploratory updates that may not yet provide a stable aggregation signal, placing \convnn at a disadvantage relative to MHSA, whose softmax weighted sum requires no additional learned weights. As the schedule reduces the learning rate toward its minimum in the final epochs, the fine-grained weight updates characteristic of low learning rates allow the aggregation kernel to converge to a precise per-channel weighting that standard uniform aggregation of KVT-attention cannot express. 

\begin{table}[t]
\centering
\normalsize
\vspace{4pt}
\caption{ImageNet-1K top-1 classification accuracy for Vision Transformer architectures. Models marked with $*$ utilize a ViT-Base backbone where the standard multi-head self-attention is replaced by the specified module. Baseline metrics for ViT, DeiT, and PVT architectures are reported as evaluated in \citet{DBLP:conf/iccv/YuanG0ZYW21}. All other baseline results are taken from their respective original publications. }
\footnotesize
\setlength{\tabcolsep}{4pt}
\begin{tabular}{lcccc}
\toprule
Model & \#Param. & Image Size & FLOPs (G) & Top-1 \\
\midrule
$*$ MHSA \cite{DBLP:conf/iclr/DosovitskiyB0WZ21}          & 86M  & $224^2$ & 35.1  & 80.9 \\
$*$ KVT ($k=9)$ \cite{DBLP:conf/eccv/WangWWLCLJ22}        & 86M  & $224^2$ & 35.1  & 80.5 \\
$*$ KVT ($k=25$) \cite{DBLP:conf/eccv/WangWWLCLJ22}       & 86M  & $224^2$ & 35.1  & 81.3 \\
$*$ KVT ($k=100$) \cite{DBLP:conf/eccv/WangWWLCLJ22}      & 86M  & $224^2$ & 35.1  & 81.0 \\
$*$ \convnn\ ($k=9$)                                      & 86M  & $224^2$ & 34.4  & 81.2 \\
$*$ Fast-\convnn\ ($k=25$)                                & 86M  & $224^2$ & 34.4  & 81.6 \\
\midrule
ViT-B/16 \cite{DBLP:conf/iclr/DosovitskiyB0WZ21}          & 86M  & $384^2$ & 55.4  & 77.9 \\
ViT-L/16 \cite{DBLP:conf/iclr/DosovitskiyB0WZ21}          & 307M & $384^2$ & 190.7 & 76.5 \\
\midrule
Swin-T  \cite{DBLP:conf/iccv/LiuL00W0LG21}                & 29M  & $224^2$ & 4.5   & 81.3 \\
Swin-S  \cite{DBLP:conf/iccv/LiuL00W0LG21}                & 50M  & $224^2$ & 8.7   & 83.0 \\
Swin-B  \cite{DBLP:conf/iccv/LiuL00W0LG21}                & 88M  & $224^2$ & 15.4  & 83.5 \\    
\midrule
DeiT-S  \cite{touvron2021training}                        & 22M  & $224^2$ & 4.5   & 79.9 \\
DeiT-B  \cite{touvron2021training}                        & 87M  & $224^2$ & 17.3  & 81.8 \\
\midrule
PVT-S  \cite{wang2021pyramid}                             & 25M  & $224^2$ & 3.8   & 79.8 \\
PVT-M  \cite{wang2021pyramid}                             & 44M  & $224^2$ & 6.7   & 81.2 \\
PVT-L  \cite{wang2021pyramid}                             & 61M  & $224^2$ & 9.8   & 81.7 \\
\midrule
CvT-13 \cite{wu2021cvt}                                   & 20M  & $224^2$ & 4.5   & 81.6 \\
CvT-21 \cite{wu2021cvt}                                   & 32M  & $224^2$ & 7.1   & 82.5 \\
\midrule    
ConViT-S  \cite{DBLP:conf/icml/dAscoliTLMBS21}            & 27M  & $224^2$ & 5.4   & 81.3 \\
ConViT-B  \cite{DBLP:conf/icml/dAscoliTLMBS21}            & 86M  & $224^2$ & 17.0  & 82.4 \\

\midrule
CeiT-T  \cite{DBLP:conf/iccv/YuanG0ZYW21}                 & 6M   & $224^2$ & 1.2   & 76.4 \\
CeiT-S  \cite{DBLP:conf/iccv/YuanG0ZYW21}                 & 24M  & $224^2$ & 4.5   & 82.0 \\
\midrule 
CoAtNet-0 \cite{DBLP:conf/nips/DaiLLT21}                  & 25M  & $224^2$ & 4.2   & 81.6 \\
CoAtNet-1 \cite{DBLP:conf/nips/DaiLLT21}                  & 42M  & $224^2$ & 8.4   & 83.3 \\
CoAtNet-2 \cite{DBLP:conf/nips/DaiLLT21}                  & 75M  & $224^2$ & 15.7  & 84.1 \\
\bottomrule
\end{tabular}
\label{tab:imagnet_vit_results}
\end{table}

\section{Conclusion}
\label{sec:conclusion}
We presented \convnn, a $k$-nearest-neighbor aggregation framework that places convolution and self-attention on a single continuum: convolution selects neighbors by spatial proximity, attention selects by feature similarity, and both are exact special cases of one operation. Beyond unifying the two, the framework yields working operations. In the convolutional setting, \emph{combining} spatial and feature-similarity selection through the hybrid branching layer (Section \ref{subsec:hybrid_branching_layer}) outperforms both a matched ResNet-50 and a pure-\convnn\ replacement, indicating that neither selection mode alone is optimal. In the attention setting, \convnn-attention surpasses multi-head self-attention and consistently beats KVT at matched neighbor counts, with its learned aggregation kernel proving especially effective at exploiting larger neighborhoods than the uniform-weight sum.

Our experiments expose several properties of neighbor selection as a design  axis. First, breadth dominates neighborhood size: ablations on $k$ reveal an asymmetry between the two operations, as larger convolutional kernels improve accuracy consistently (e.g., kernel size $1 \rightarrow 2$ or $3$), whereas increasing $k$ in \convnn\ yields only modest gains that plateau beyond a small value. Second, candidate breadth matters more than candidate sparsity: when employing a global receptive field, drawing neighbors from the full candidate pool is more effective than from a randomly or spatially subsampled subset, and the more candidates \convnn\ can select from, the better. This effect is attenuated within the hybrid branching layer, where the parallel convolution branch counterbalances the sparsity of a subsampled global selection. Third, the \convnn\ branch acts as an implicit regularizer in convolutional architectures
introducing a global, feature-based pathway alongside local convolution forces  the network to balance the two and discourages over-reliance on either alone.

A further finding concerns training dynamics. \convnn\ exhibits a distinctive late-stage convergence pattern, trailing multi-head attention through the early  and middle epochs before overtaking it in the final stages. We attribute this to the learned aggregation kernel, which encodes a form of neighbor-channel importance that develops only under the precise, low-magnitude gradient updates of the low-learning-rate schedule; it suggests, more broadly, that the advantage  of learnable over fixed-uniform aggregation is most visible under low-learning-rate
fine-tuning.

We do not target architecture-level state of the art. Stronger results from bespoke hybrid designs (e.g., CoAtNet, ConvNeXt) reflect orthogonal advances in macro-architecture, whereas \convnn\ operates at the level of the underlying 
aggregation operation; our claims are accordingly framed against matched baselines that isolate the effect of \emph{how} neighbors are selected. Two questions remain open. First, while we characterize both endpoints and several discrete configurations of the spectrum, whether an \emph{intermediate} selection criterion can be tuned to outperform both endpoints in general is unresolved, and is the most natural direction the framework opens. Second, the schedule-dependence of the late-stage convergence behavior—learnable aggregation paying off only under low-learning-rate fine-tuning—merits dedicated study, as it bears on how such operations should be trained in practice.
\section{Acknowledgments}
This work used Jetstream2 \cite{10.1145/3437359.3465565} at Indiana University through allocation SEE240007 from the Advanced Cyberinfrastructure Coordination Ecosystem: Services \& Support (ACCESS) \cite{Boerneretal2023} program, which is supported by National Science Foundation grants \#2138259, \#2138286, \#2138307, \#2137603, and \#2138296.

This research is supported by the National Artificial Intelligence Research Resource (NAIRR) Pilot allocation NAIRR260070 and the Jetstream2 cloud resource supported by the National Science Foundation (award NSF-OAC 2005506) at Indiana University.


{
    \small
    \bibliographystyle{ieeenat_fullname}
    \bibliography{main}

@String(IJCV = {Int. J. Comput. Vis.})

@String(CVPR= {IEEE Conf. Comput. Vis. Pattern Recog.})

@String(ICCV= {Int. Conf. Comput. Vis.})

@String(ECCV= {Eur. Conf. Comput. Vis.})

@String(ICLR = {Int. Conf. Learn. Represent.})

@String(IJCV  = {IJCV})

@String(CVPR  = {CVPR})

@String(ICCV  = {ICCV})

@String(ECCV  = {ECCV})

@String(ICLR  = {ICLR})

@article{vaswani2017attention,
  title={Attention is all you need},
  author={Vaswani, Ashish and Shazeer, Noam and Parmar, Niki and Uszkoreit, Jakob and Jones, Llion and Gomez, Aidan N and Kaiser, {\L}ukasz and Polosukhin, Illia},
  journal={Advances in neural information processing systems},
  volume={30},
  year={2017}
}

@inproceedings{he2016deep,
  title={Deep residual learning for image recognition},
  author={He, Kaiming and Zhang, Xiangyu and Ren, Shaoqing and Sun, Jian},
  booktitle={Proceedings of the IEEE conference on computer vision and pattern recognition},
  pages={770--778},
  year={2016}
}

@inproceedings{DBLP:journals/corr/SimonyanZ14a,
  author       = {Karen Simonyan and
                  Andrew Zisserman},
  editor       = {Yoshua Bengio and
                  Yann LeCun},
  title        = {Very Deep Convolutional Networks for Large-Scale Image Recognition},
  booktitle    = {3rd International Conference on Learning Representations, {ICLR} 2015,
                  San Diego, CA, USA, May 7-9, 2015, Conference Track Proceedings},
  year         = {2015},
  url          = {http://arxiv.org/abs/1409.1556},
  bibsource    = {dblp computer science bibliography, https://dblp.org}
}

@inproceedings{DBLP:conf/iclr/DosovitskiyB0WZ21,
  author       = {Alexey Dosovitskiy and
                  Lucas Beyer and
                  Alexander Kolesnikov and
                  Dirk Weissenborn and
                  Xiaohua Zhai and
                  Thomas Unterthiner and
                  Mostafa Dehghani and
                  Matthias Minderer and
                  Georg Heigold and
                  Sylvain Gelly and
                  Jakob Uszkoreit and
                  Neil Houlsby},
  title        = {An Image is Worth 16x16 Words: Transformers for Image Recognition
                  at Scale},
  booktitle    = {9th International Conference on Learning Representations, {ICLR} 2021,
                  Virtual Event, Austria, May 3-7, 2021},
  publisher    = {OpenReview.net},
  year         = {2021},
  url          = {https://openreview.net/forum?id=YicbFdNTTy},
  bibsource    = {dblp computer science bibliography, https://dblp.org}
}

@article{DBLP:journals/cacm/KrizhevskySH17,
  author       = {Alex Krizhevsky and
                  Ilya Sutskever and
                  Geoffrey E. Hinton},
  title        = {ImageNet classification with deep convolutional neural networks},
  journal      = {Commun. {ACM}},
  volume       = {60},
  number       = {6},
  pages        = {84--90},
  year         = {2017},
  url          = {https://doi.org/10.1145/3065386},
  doi          = {10.1145/3065386},
  bibsource    = {dblp computer science bibliography, https://dblp.org}
}

@inproceedings{DBLP:conf/nips/Plotz018,
  author       = {Tobias Pl{\"{o}}tz and
                  Stefan Roth},
  editor       = {Samy Bengio and
                  Hanna M. Wallach and
                  Hugo Larochelle and
                  Kristen Grauman and
                  Nicol{\`{o}} Cesa{-}Bianchi and
                  Roman Garnett},
  title        = {Neural Nearest Neighbors Networks},
  booktitle    = {Advances in Neural Information Processing Systems 31: Annual Conference
                  on Neural Information Processing Systems 2018, NeurIPS 2018, December
                  3-8, 2018, Montr{\'{e}}al, Canada},
  pages        = {1095--1106},
  year         = {2018},
  url          = {https://proceedings.neurips.cc/paper/2018/hash/f0e52b27a7a5d6a1a87373dffa53dbe5-Abstract.html},
  bibsource    = {dblp computer science bibliography, https://dblp.org}
}

@inproceedings{DBLP:conf/cvpr/0004GGH18,
  author       = {Xiaolong Wang and
                  Ross B. Girshick and
                  Abhinav Gupta and
                  Kaiming He},
  title        = {Non-Local Neural Networks},
  booktitle    = {2018 {IEEE} Conference on Computer Vision and Pattern Recognition,
                  {CVPR} 2018, Salt Lake City, UT, USA, June 18-22, 2018},
  pages        = {7794--7803},
  publisher    = {Computer Vision Foundation / {IEEE} Computer Society},
  year         = {2018},
  url          = {http://openaccess.thecvf.com/content\_cvpr\_2018/html/Wang\_Non-Local\_Neural\_Networks\_CVPR\_2018\_paper.html},
  doi          = {10.1109/CVPR.2018.00813},
  bibsource    = {dblp computer science bibliography, https://dblp.org}
}

@inproceedings{DBLP:conf/iclr/CordonnierLJ20,
  author       = {Jean{-}Baptiste Cordonnier and
                  Andreas Loukas and
                  Martin Jaggi},
  title        = {On the Relationship between Self-Attention and Convolutional Layers},
  booktitle    = {8th International Conference on Learning Representations, {ICLR} 2020,
                  Addis Ababa, Ethiopia, April 26-30, 2020},
  publisher    = {OpenReview.net},
  year         = {2020},
  url          = {https://openreview.net/forum?id=HJlnC1rKPB},
  bibsource    = {dblp computer science bibliography, https://dblp.org}
}

@inproceedings{DBLP:conf/icml/dAscoliTLMBS21,
  author       = {St{\'{e}}phane d'Ascoli and
                  Hugo Touvron and
                  Matthew L. Leavitt and
                  Ari S. Morcos and
                  Giulio Biroli and
                  Levent Sagun},
  editor       = {Marina Meila and
                  Tong Zhang},
  title        = {ConViT: Improving Vision Transformers with Soft Convolutional Inductive
                  Biases},
  booktitle    = {Proceedings of the 38th International Conference on Machine Learning,
                  {ICML} 2021, 18-24 July 2021, Virtual Event},
  series       = {Proceedings of Machine Learning Research},
  volume       = {139},
  pages        = {2286--2296},
  publisher    = {{PMLR}},
  year         = {2021},
  url          = {http://proceedings.mlr.press/v139/d-ascoli21a.html},
  bibsource    = {dblp computer science bibliography, https://dblp.org}
}

@inproceedings{DBLP:conf/iccv/YuanG0ZYW21,
  author       = {Kun Yuan and
                  Shaopeng Guo and
                  Ziwei Liu and
                  Aojun Zhou and
                  Fengwei Yu and
                  Wei Wu},
  title        = {Incorporating Convolution Designs into Visual Transformers},
  booktitle    = {2021 {IEEE/CVF} International Conference on Computer Vision, {ICCV}
                  2021, Montreal, QC, Canada, October 10-17, 2021},
  pages        = {559--568},
  publisher    = {{IEEE}},
  year         = {2021},
  url          = {https://doi.org/10.1109/ICCV48922.2021.00062},
  doi          = {10.1109/ICCV48922.2021.00062},
  bibsource    = {dblp computer science bibliography, https://dblp.org}
}

@inproceedings{DBLP:conf/iccv/LiuL00W0LG21,
  author       = {Ze Liu and
                  Yutong Lin and
                  Yue Cao and
                  Han Hu and
                  Yixuan Wei and
                  Zheng Zhang and
                  Stephen Lin and
                  Baining Guo},
  title        = {Swin Transformer: Hierarchical Vision Transformer using Shifted Windows},
  booktitle    = {2021 {IEEE/CVF} International Conference on Computer Vision, {ICCV}
                  2021, Montreal, QC, Canada, October 10-17, 2021},
  pages        = {9992--10002},
  publisher    = {{IEEE}},
  year         = {2021},
  url          = {https://doi.org/10.1109/ICCV48922.2021.00986},
  doi          = {10.1109/ICCV48922.2021.00986},
  bibsource    = {dblp computer science bibliography, https://dblp.org}
}

@article{DBLP:journals/pami/HuSASW20,
  author       = {Jie Hu and
                  Li Shen and
                  Samuel Albanie and
                  Gang Sun and
                  Enhua Wu},
  title        = {Squeeze-and-Excitation Networks},
  journal      = {{IEEE} Trans. Pattern Anal. Mach. Intell.},
  volume       = {42},
  number       = {8},
  pages        = {2011--2023},
  year         = {2020},
  url          = {https://doi.org/10.1109/TPAMI.2019.2913372},
  doi          = {10.1109/TPAMI.2019.2913372},
  bibsource    = {dblp computer science bibliography, https://dblp.org}
}

@inproceedings{DBLP:conf/iccv/BelloZLVS19,
  author       = {Irwan Bello and
                  Barret Zoph and
                  Quoc Le and
                  Ashish Vaswani and
                  Jonathon Shlens},
  title        = {Attention Augmented Convolutional Networks},
  booktitle    = {2019 {IEEE/CVF} International Conference on Computer Vision, {ICCV}
                  2019, Seoul, Korea (South), October 27 - November 2, 2019},
  pages        = {3285--3294},
  publisher    = {{IEEE}},
  year         = {2019},
  url          = {https://doi.org/10.1109/ICCV.2019.00338},
  doi          = {10.1109/ICCV.2019.00338},
  bibsource    = {dblp computer science bibliography, https://dblp.org}
}

@inproceedings{DBLP:conf/eccv/WangWWLCLJ22,
  author       = {Pichao Wang and
                  Xue Wang and
                  Fan Wang and
                  Ming Lin and
                  Shuning Chang and
                  Hao Li and
                  Rong Jin},
  editor       = {Shai Avidan and
                  Gabriel J. Brostow and
                  Moustapha Ciss{\'{e}} and
                  Giovanni Maria Farinella and
                  Tal Hassner},
  title        = {{KVT:} k-NN Attention for Boosting Vision Transformers},
  booktitle    = {Computer Vision - {ECCV} 2022: 17th European Conference, Tel Aviv,
                  Israel, October 23-27, 2022, Proceedings, Part {XXIV}},
  series       = {Lecture Notes in Computer Science},
  volume       = {13684},
  pages        = {285--302},
  publisher    = {Springer},
  year         = {2022},
  url          = {https://doi.org/10.1007/978-3-031-20053-3\_17},
  doi          = {10.1007/978-3-031-20053-3\_17},
  bibsource    = {dblp computer science bibliography, https://dblp.org}
}

@inproceedings{DBLP:conf/nips/DaiLLT21,
  author       = {Zihang Dai and
                  Hanxiao Liu and
                  Quoc V. Le and
                  Mingxing Tan},
  editor       = {Marc'Aurelio Ranzato and
                  Alina Beygelzimer and
                  Yann N. Dauphin and
                  Percy Liang and
                  Jennifer Wortman Vaughan},
  title        = {CoAtNet: Marrying Convolution and Attention for All Data Sizes},
  booktitle    = {Advances in Neural Information Processing Systems 34: Annual Conference
                  on Neural Information Processing Systems 2021, NeurIPS 2021, December
                  6-14, 2021, virtual},
  pages        = {3965--3977},
  year         = {2021},
  url          = {https://proceedings.neurips.cc/paper/2021/hash/20568692db622456cc42a2e853ca21f8-Abstract.html},
  bibsource    = {dblp computer science bibliography, https://dblp.org}
}

@inproceedings{DBLP:conf/nips/Bertsch0NG23,
  author       = {Amanda Bertsch and
                  Uri Alon and
                  Graham Neubig and
                  Matthew Gormley},
  editor       = {Alice Oh and
                  Tristan Naumann and
                  Amir Globerson and
                  Kate Saenko and
                  Moritz Hardt and
                  Sergey Levine},
  title        = {Unlimiformer: Long-Range Transformers with Unlimited Length Input},
  booktitle    = {Advances in Neural Information Processing Systems 36: Annual Conference
                  on Neural Information Processing Systems 2023, NeurIPS 2023, New Orleans,
                  LA, USA, December 10 - 16, 2023},
  year         = {2023},
  url          = {http://papers.nips.cc/paper\_files/paper/2023/hash/6f9806a5adc72b5b834b27e4c7c0df9b-Abstract-Conference.html},
  bibsource    = {dblp computer science bibliography, https://dblp.org}
}

@article{roy-etal-2021-efficient,
    title = "Efficient Content-Based Sparse Attention with Routing Transformers",
    author = "Roy, Aurko  and
      Saffar, Mohammad  and
      Vaswani, Ashish  and
      Grangier, David",
    editor = "Roark, Brian  and
      Nenkova, Ani",
    journal = "Transactions of the Association for Computational Linguistics",
    volume = "9",
    year = "2021",
    address = "Cambridge, MA",
    publisher = "MIT Press",
    url = "https://aclanthology.org/2021.tacl-1.4/",
    doi = "10.1162/tacl_a_00353",
    pages = "53--68"
}

@inproceedings{szegedy2015going,
  title     = {Going Deeper with Convolutions},
  author    = {Szegedy, Christian and Liu, Wei and Jia, Yangqing and Sermanet, Pierre and Reed, Scott and Anguelov, Dragomir and Erhan, Dumitru and Vanhoucke, Vincent and Rabinovich, Andrew},
  booktitle = {Proceedings of the IEEE Conference on Computer Vision and Pattern Recognition (CVPR)},
  year      = {2015},
  pages     = {1--9},
  doi       = {10.1109/CVPR.2015.7298594}
}

@inproceedings{peng2021conformer,
  title={Conformer: Local features coupling global representations for visual recognition},
  author={Peng, Zhiliang and Huang, Wei and Gu, Shanzhi and Xie, Lingxi and Wang, Yaowei and Jiao, Jianbin and Ye, Qixiang},
  booktitle={Proceedings of the IEEE/CVF international conference on computer vision},
  pages={367--376},
  year={2021}
}

@inproceedings{chen2022regionvit,
  title     = {RegionViT: Regional-to-Local Attention for Vision Transformers},
  author    = {Chen, Chun-Fu and Panda, Rameswar and Fan, Quanfu},
  booktitle = {Proceedings of the International Conference on Learning Representations (ICLR)},
  year      = {2022},
  note      = {Poster},
  url       = {https://arxiv.org/abs/2106.02689},
}

@inproceedings{child2019generating,
  title     = {Generating Long Sequences with Sparse Transformers},
  author    = {Child, Rewon and Gray, Scott and Radford, Alec and Sutskever, Ilya},
  booktitle = {Proceedings of the International Conference on Learning Representations (ICLR)},
  year      = {2019},
  url       = {https://arxiv.org/abs/1904.10509}
}

@inproceedings{Boerneretal2023,
  author    = {Boerner, Timothy J. and Deems, Stephen and Furlani, Thomas R. and Knuth, Shelley L. and Towns, John},
  title     = {ACCESS: Advancing Innovation: NSF's Advanced Cyberinfrastructure Coordination Ecosystem: Services \& Support},
  booktitle = {Practice and Experience in Advanced Research Computing (PEARC '23)},
  year      = {2023},
  publisher = {Association for Computing Machinery},
  address   = {New York, NY, USA},
  pages     = {173--176},
  doi       = {10.1145/3569951.3597559}
}

@inproceedings{touvron2021training,
  title={Training data-efficient image transformers \& distillation through attention},
  author={Touvron, Hugo and Cord, Matthieu and Douze, Matthijs and Massa, Francisco and Sablayrolles, Alexandre and J{\'e}gou, Herv{\'e}},
  booktitle={International conference on machine learning},
  pages={10347--10357},
  year={2021},
  organization={PMLR}
}

@inproceedings{katharopoulos2020transformers,
  title={Transformers are rnns: Fast autoregressive transformers with linear attention},
  author={Katharopoulos, Angelos and Vyas, Apoorv and Pappas, Nikolaos and Fleuret, Fran{\c{c}}ois},
  booktitle={International conference on machine learning},
  pages={5156--5165},
  year={2020},
  organization={PMLR}
}

@article{loshchilov2017decoupled,
  title={Decoupled weight decay regularization},
  author={Loshchilov, Ilya and Hutter, Frank},
  journal={arXiv preprint arXiv:1711.05101},
  year={2017}
}

@article{hugger2024towards,
  title={Towards noise contrastive estimation with soft targets for conditional models},
  author={Hugger, Johannes and Uhlmann, Virginie},
  journal={arXiv preprint arXiv:2404.14076},
  year={2024}
}

@article{loshchilov2016sgdr,
  title={Sgdr: Stochastic gradient descent with warm restarts},
  author={Loshchilov, Ilya and Hutter, Frank},
  journal={arXiv preprint arXiv:1608.03983},
  year={2016}
}

@article{ILSVRC15,
Author = {Olga Russakovsky and Jia Deng and Hao Su and Jonathan Krause and Sanjeev Satheesh and Sean Ma and Zhiheng Huang and Andrej Karpathy and Aditya Khosla and Michael Bernstein and Alexander C. Berg and Li Fei-Fei},
Title = {{ImageNet Large Scale Visual Recognition Challenge}},
Year = {2015},
journal   = {International Journal of Computer Vision (IJCV)},
doi = {10.1007/s11263-015-0816-y},
volume={115},
number={3},
pages={211-252}
}

@inproceedings{yun2019cutmix,
  title={Cutmix: Regularization strategy to train strong classifiers with localizable features},
  author={Yun, Sangdoo and Han, Dongyoon and Oh, Seong Joon and Chun, Sanghyuk and Choe, Junsuk and Yoo, Youngjoon},
  booktitle={Proceedings of the IEEE/CVF international conference on computer vision},
  pages={6023--6032},
  year={2019}
}

@article{zhang2017mixup,
  title={mixup: Beyond empirical risk minimization},
  author={Zhang, Hongyi and Cisse, Moustapha and Dauphin, Yann N and Lopez-Paz, David},
  journal={arXiv preprint arXiv:1710.09412},
  year={2017}
}

@inproceedings{cubuk2020randaugment,
  title={Randaugment: Practical automated data augmentation with a reduced search space},
  author={Cubuk, Ekin D and Zoph, Barret and Shlens, Jonathon and Le, Quoc V},
  booktitle={Proceedings of the IEEE/CVF conference on computer vision and pattern recognition workshops},
  pages={702--703},
  year={2020}
}

@article{zaheer2020big,
  title={Big bird: Transformers for longer sequences},
  author={Zaheer, Manzil and Guruganesh, Guru and Dubey, Kumar Avinava and Ainslie, Joshua and Alberti, Chris and Ontanon, Santiago and Pham, Philip and Ravula, Anirudh and Wang, Qifan and Yang, Li and others},
  journal={Advances in neural information processing systems},
  volume={33},
  pages={17283--17297},
  year={2020}
}

@inproceedings{10.1145/3315508.3329973,
author = {Tillet, Philippe and Kung, H. T. and Cox, David},
title = {Triton: an intermediate language and compiler for tiled neural network computations},
year = {2019},
isbn = {9781450367196},
publisher = {Association for Computing Machinery},
address = {New York, NY, USA},
url = {https://doi.org/10.1145/3315508.3329973},
doi = {10.1145/3315508.3329973},
booktitle = {Proceedings of the 3rd ACM SIGPLAN International Workshop on Machine Learning and Programming Languages},
pages = {10–19},
numpages = {10},
location = {Phoenix, AZ, USA},
series = {MAPL 2019}
}

@article{han2021connection,
  title={On the connection between local attention and dynamic depth-wise convolution},
  author={Han, Qi and Fan, Zejia and Dai, Qi and Sun, Lei and Cheng, Ming-Ming and Liu, Jiaying and Wang, Jingdong},
  journal={arXiv preprint arXiv:2106.04263},
  year={2021}
}

@inproceedings{10.1145/3437359.3465565,
author = {Hancock, David Y. and Fischer, Jeremy and Lowe, John Michael and Snapp-Childs, Winona and Pierce, Marlon and Marru, Suresh and Coulter, J. Eric and Vaughn, Matthew and Beck, Brian and Merchant, Nirav and Skidmore, Edwin and Jacobs, Gwen},
title = {Jetstream2: Accelerating cloud computing via Jetstream},
year = {2021},
isbn = {9781450382922},
publisher = {Association for Computing Machinery},
address = {New York, NY, USA},
url = {https://doi.org/10.1145/3437359.3465565},
doi = {10.1145/3437359.3465565},
booktitle = {Practice and Experience in Advanced Research Computing 2021: Evolution Across All Dimensions},
articleno = {11},
numpages = {8},
location = {Boston, MA, USA},
series = {PEARC '21}
}

@inproceedings{pan2022integration,
  title={On the integration of self-attention and convolution},
  author={Pan, Xuran and Ge, Chunjiang and Lu, Rui and Song, Shiji and Chen, Guanfu and Huang, Zeyi and Huang, Gao},
  booktitle={Proceedings of the IEEE/CVF conference on computer vision and pattern recognition},
  pages={815--825},
  year={2022}
}

@article{zhang2023vision,
  title={Vision Big Bird: Random Sparsification for Full Attention},
  author={Zhang, Zhemin and Gong, Xun},
  journal={arXiv preprint arXiv:2311.05988},
  year={2023}
}

@inproceedings{liu2022convnet,
  title={A convnet for the 2020s},
  author={Liu, Zhuang and Mao, Hanzi and Wu, Chao-Yuan and Feichtenhofer, Christoph and Darrell, Trevor and Xie, Saining},
  booktitle={Proceedings of the IEEE/CVF conference on computer vision and pattern recognition},
  pages={11976--11986},
  year={2022}
}

@inproceedings{wu2021cvt,
  title={Cvt: Introducing convolutions to vision transformers},
  author={Wu, Haiping and Xiao, Bin and Codella, Noel and Liu, Mengchen and Dai, Xiyang and Yuan, Lu and Zhang, Lei},
  booktitle={Proceedings of the IEEE/CVF international conference on computer vision},
  pages={22--31},
  year={2021}
}

@inproceedings{wang2021pyramid,
  title={Pyramid vision transformer: A versatile backbone for dense prediction without convolutions},
  author={Wang, Wenhai and Xie, Enze and Li, Xiang and Fan, Deng-Ping and Song, Kaitao and Liang, Ding and Lu, Tong and Luo, Ping and Shao, Ling},
  booktitle={Proceedings of the IEEE/CVF international conference on computer vision},
  pages={568--578},
  year={2021}
}

@inproceedings{radosavovic2020designing,
  title={Designing network design spaces},
  author={Radosavovic, Ilija and Kosaraju, Raj Prateek and Girshick, Ross and He, Kaiming and Doll{\'a}r, Piotr},
  booktitle={Proceedings of the IEEE/CVF conference on computer vision and pattern recognition},
  pages={10428--10436},
  year={2020}
}

@inproceedings{tan2019efficientnet,
  title={Efficientnet: Rethinking model scaling for convolutional neural networks},
  author={Tan, Mingxing and Le, Quoc},
  booktitle={International conference on machine learning},
  pages={6105--6114},
  year={2019},
  organization={PMLR}
}
}

\clearpage
\newpage
\setcounter{page}{1}
\onecolumn

%
%

\newcommand{\restatetheorem}[3]{%
  \par\medskip\noindent\textbf{Theorem~\ref{#1} (#2).}\;\itshape #3\upshape\par\medskip}

\begin{center}
    \Large
    \textbf{\thetitle}\\
    \vspace{0.5em}Supplementary Material \\
    \vspace{1.0em}
\end{center}

\section{Convolution and Attention Formulation}
\label{sec:supp_formulation}

This section fixes the notation and recalls the standard definitions of
convolution, self-attention, and multi-head attention that the proofs in
Sections~\ref{proof:conv_connection} and~\ref{proof:attention_generalization}
reduce \convnn\ to.

\subsection{Convolution}
We consider two variants of convolution, standard and depthwise. For the 2D
case, let $\mathbf{X} \in \mathbb{R}^{H \times W \times C}$ denote an input
feature map with spatial dimensions $H \times W$ and $C$ channels; for the 1D
case, let $\mathbf{x} \in \mathbb{R}^{N \times C}$ denote an input with sequence
length $N$ and $C$ channels. In both cases, $R$ denotes the spatial kernel size
and $C'$ the number of output channels.

\subsubsection{Standard Convolution}
\label{subsec:std_conv}
Standard convolution applies a learned kernel
$W \in \mathbb{R}^{R \times R \times C \times C'}$ over the full spatial and
channel extent of the input. For a 2D input,
\begin{equation}
    \operatorname{Conv}(W, \mathbf{X})_{(i,j,c')}
    = \sum_{p=1}^{R}\sum_{q=1}^{R}\sum_{c=1}^{C}
      W_{p,q,c,c'} \cdot \mathbf{X}_{i+p,\, j+q,\, c}.
    \label{eq:std_conv_2d}
\end{equation}
For a 1D input with kernel $W \in \mathbb{R}^{R \times C \times C'}$, the spatial
summation reduces to a single dimension,
\begin{equation}
    \operatorname{Conv}(W, \mathbf{x})_{(i,c')}
    = \sum_{p=1}^{R}\sum_{c=1}^{C}
      W_{p,c,c'} \cdot \mathbf{x}_{i+p,\, c}.
    \label{eq:std_conv_1d}
\end{equation}

\subsubsection{Depthwise Convolution}
\label{subsec:dw_conv}
Depthwise convolution applies a separate spatial kernel to each input channel
independently, with no cross-channel mixing. For a 2D input with per-channel
kernel $W \in \mathbb{R}^{R \times R \times C}$,
\begin{equation}
    \operatorname{DepthwiseConv}(W, \mathbf{X})_{(i,j,c)}
    = \sum_{p=1}^{R}\sum_{q=1}^{R}
      W_{p,q,c} \cdot \mathbf{X}_{i+p,\, j+q,\, c}.
    \label{eq:dw_conv_2d}
\end{equation}
For a 1D input with kernel $W \in \mathbb{R}^{R \times C}$, the per-channel
spatial filtering reduces to
\begin{equation}
    \operatorname{DepthwiseConv}(W, \mathbf{x})_{(i,c)}
    = \sum_{p=1}^{R}
      W_{p,c} \cdot \mathbf{x}_{i+p,\, c}.
    \label{eq:dw_conv_1d}
\end{equation}

\subsection{Self-Attention}
\label{subsec:self_attention}
Given an input sequence $X \in \mathbb{R}^{N \times d_{\text{model}}}$, queries,
keys, and values are obtained via learned linear projections
$Q = f_Q(X) = XW^Q$, $K = f_K(X) = XW^K$, and $V = f_V(X) = XW^V$, with
$W^Q, W^K, W^V \in \mathbb{R}^{d_{\text{model}} \times d_k}$, where
$d_{\text{model}}$ is the input feature dimension and $d_k$ is the projected
embedding dimension. The attention output is
\begin{equation}
    \operatorname{Attention}(Q, K, V)
    = \operatorname{softmax}\!\left(\frac{QK^\top}{\sqrt{d_k}}\right)V,
    \label{eq:self_attention}
\end{equation}
where the $\operatorname{softmax}$ is applied row-wise to the scaled score
matrix $QK^\top / \sqrt{d_k} \in \mathbb{R}^{N \times N}$, converting each row
into a probability distribution over positions,
\begin{equation}
    \operatorname{softmax}(z)_j = \frac{\exp(z_j)}{\sum_{j'=1}^{N} \exp(z_{j'})}.
    \label{eq:softmax}
\end{equation}
Thus each output position $i$ attends to all $N$ positions with non-negative
weights that sum to one. The scaling factor $1/\sqrt{d_k}$ in
Equation~\eqref{eq:self_attention} prevents the dot products from growing large
in magnitude as $d_k$ increases, which would otherwise concentrate the softmax
distribution toward a single position and produce vanishing gradients elsewhere.

In elementwise form, the query, key, and value vectors for position $i$ are
$q_i = W^Q x_i$, $k_j = W^K x_j$, and $v_j = W^V x_j$. The attention weight and
output at position $i$ are
\begin{equation}
    a_{i,j}
    = \frac{\exp(q_i^\top k_j / \sqrt{d_k})}
           {\sum_{j'=1}^{N} \exp(q_i^\top k_{j'} / \sqrt{d_k})},
    \qquad
    y_i = \sum_{j=1}^{N} a_{i,j}\, v_j.
    \label{eq:self_attention_partition}
\end{equation}

\paragraph{Causal (Masked) Self-Attention.}
\label{par:causal_self_attention}
Causal (masked) self-attention is used in autoregressive sequence modeling,
where each position $i$ may attend only to positions $j \leq i$, i.e.\ to itself
and all preceding elements. This constraint prevents the model from accessing
future positions during training, ensuring that the prediction at position $i$
depends only on the observed context $(x_1, \dots, x_i)$. The causal attention
weight restricts the softmax normalization to the valid context window,
\begin{equation}
    a_{i,j}^{\text{causal}}
    = \frac{\exp(q_i^\top k_j / \sqrt{d_k})}
           {\sum_{j'=1}^{i} \exp(q_i^\top k_{j'} / \sqrt{d_k})},
    \qquad j \leq i,
    \label{eq:causal_attention}
\end{equation}
so that the output at each position aggregates only over preceding and current
values,
\begin{equation}
    y_i = \sum_{j=1}^{i} a_{i,j}^{\text{causal}}\, v_j.
    \label{eq:causal_output}
\end{equation}
In practice, causal masking is implemented by adding a mask matrix
$M \in \{0, -\infty\}^{N \times N}$ to the pre-softmax scores, with $M_{ij} = 0$
if $j \leq i$ and $M_{ij} = -\infty$ otherwise,
\begin{equation}
    \operatorname{CausalAttention}(Q, K, V)
    = \operatorname{softmax}\!\left(
        \frac{QK^\top}{\sqrt{d_k}} + M
      \right) V.
    \label{eq:causal_attention_matrix}
\end{equation}
Since $\exp(-\infty) = 0$, the softmax assigns exactly zero weight to all future
positions, zeroing out the strictly-upper-triangular entries of the attention
matrix. The operation is thus fully autoregressive while remaining
parallelizable across all positions during training, unlike recurrent
architectures that must process tokens sequentially.

\subsection{Multi-Head Attention}
\label{subsec:multihead_attention}
Multi-head attention runs $H$ attention operations in parallel over learned
subspaces of the projected representations,
\begin{align}
    \operatorname{MultiHead}(Q, K, V)
    &= \operatorname{Concat}(\text{head}_1, \dots, \text{head}_H)\,W^O,
    \label{eq:multihead_attention} \\
    \text{head}_h
    &= \operatorname{Attention}(QW_h^Q,\; KW_h^K,\; VW_h^V),
    \label{eq:multihead_head}
\end{align}
where $W_h^Q, W_h^K, W_h^V \in \mathbb{R}^{d_\text{model} \times d_k/H}$ are
per-head projection matrices and $W^O \in \mathbb{R}^{d_k \times d_{\text{model}}}$
is the output projection.

In practice, multi-head attention is implemented not by running $H$ independent
operations sequentially, but by introducing a head dimension directly into the
projected tensors. Given $X \in \mathbb{R}^{N \times d_{\text{model}}}$, the full
projections $Q, K, V \in \mathbb{R}^{N \times d_k}$ are computed and then
reshaped by partitioning the $d_k$ embedding dimension evenly across $H$ heads,
\begin{equation}
    Q_H, K_H, V_H \in \mathbb{R}^{H \times N \times d_k/H},
    \label{eq:multihead_reshape}
\end{equation}
where $d_k$ must be divisible by $H$. Each head $h$ operates on a distinct
$d_k/H$-dimensional subspace, and the attention computation proceeds in parallel
across all heads with the per-head scaling $1/\sqrt{d_k/H}$,
\begin{equation}
    \operatorname{MultiHead}(Q_H, K_H, V_H)
    = \operatorname{softmax}\!\left(
        \frac{Q_H K_H^\top}{\sqrt{d_k / H}}
      \right) V_H,
    \label{eq:multihead_attention_split}
\end{equation}
where $\operatorname{softmax}(Q_H K_H^\top / \sqrt{d_k/H}) \in \mathbb{R}^{H \times N \times N}$
and $V_H \in \mathbb{R}^{H \times N \times d_k/H}$, yielding an output in
$\mathbb{R}^{H \times N \times d_k/H}$. The head dimension is then collapsed by
concatenation along the last axis, recovering a tensor of shape
$\mathbb{R}^{N \times d_k}$, which is finally projected by
$W^O \in \mathbb{R}^{d_k \times d_{\text{model}}}$,
\begin{equation}
    Y = \operatorname{Concat}_H\!\left(
        \operatorname{softmax}\!\left(
            \frac{Q_H K_H^\top}{\sqrt{d_k/H}}
        \right) V_H
    \right) W^O \in \mathbb{R}^{N \times d_{\text{model}}}.
    \label{eq:multihead_output}
\end{equation}
This is mathematically equivalent to the concatenation of $H$ independent
attention heads in Equation~\eqref{eq:multihead_attention}, but is more efficient
as it replaces sequential head computation with batched matrix multiplication
across the head dimension.

\section{Proofs for the Connection to Convolution}
\label{proof:conv_connection}

\subsection{Proof of Theorem~\ref{thm:conv_equivalence}}
\label{proof:conv_equivalence}

\restatetheorem{thm:conv_equivalence}{Equivalence to Convolutional Operations}{%
Let $X$ be an input tensor augmented with normalized spatial coordinates. If the
query and key projections are restricted to the raw spatial coordinates, the
similarity metric is the exponential of the negative squared Euclidean distance 
$S_{ij} = \exp(-\|p_i - p_j\|_2^2)$, and neighbor selection retrieves $K = R^{d}$
neighbors (with $R$ an odd kernel size and $d$ the spatial dimension), then
\convnn\ under uniform weighting $\rho(\mathbf{z}) = \mathbf{1}_K$ exactly
recovers standard and depthwise convolution at all interior spatial positions.}

\begin{proof}
We argue constructively for $d=1$ and $d=2$, showing that under the stated
construction the \convnn\ aggregation output is index-by-index identical to
standard and depthwise convolution at every interior position. We treat the 1D
case in full and then reduce the 2D case to it.

\medskip
\noindent\emph{1D construction.}
Given $X \in \mathbb{R}^{N \times C}$, append the normalized coordinate
$p_i = i/(N-1) \in [0,1]$ along the channel dimension, yielding
$X_{\text{pos}} \in \mathbb{R}^{N \times (C+1)}$, and define
\begin{equation}
    Q = X_{\text{pos}}[:, -1{:}], \quad
    K = X_{\text{pos}}[:, -1{:}], \quad
    V = X,
    \qquad Q, K \in \mathbb{R}^{N \times 1},\quad V \in \mathbb{R}^{N \times C}.
    \label{eq:conv1d_proj}
\end{equation}

\emph{Similarity.} The similarity matrix is the exponential of the negative squared
Euclidean distance between positional coordinates,
\begin{equation}
    S_{ij} = \exp(-\|p_i - p_j\|_2^2)
           = \exp\left(-\left(\frac{i - j}{N-1}\right)^2\right),
    \qquad S \in \mathbb{R}^{N \times N}.
    \label{eq:conv1d_sim}
\end{equation}
Because $S_{ij}$ is strictly decreasing in $|i - j|$, positions nearer in index
receive strictly higher scores, so the ordering induced by $S$ along each row is
exactly the ordering by index distance.

\emph{Neighbor selection.} Applying $\argkmax_k$ with $K = R$ to each row of $S$
therefore selects the $R$ positions nearest in index. For an interior position
$\lfloor R/2 \rfloor \leq i \leq N - 1 - \lfloor R/2 \rfloor$, this is the
symmetric centered window
\begin{equation}
    I_i = \left\{\, i - \left\lfloor\tfrac{R}{2}\right\rfloor,\;
                 \dots,\; i,\; \dots,\;
                 i + \left\lfloor\tfrac{R}{2}\right\rfloor \,\right\}
        \in \{0, \dots, N-1\}^{R},
    \label{eq:conv1d_window}
\end{equation}
with $i$ at the center, matching the standard centered-convolution convention.

\emph{Modulation.} With uniform weights $\rho(\mathbf{z}) = \mathbf{1}_K$, the
neighborhood matrix for position $i$ is
\begin{equation}
    X_{\text{nn},i} = V[I_i, :] \in \mathbb{R}^{R \times C}.
    \label{eq:conv1d_nn}
\end{equation}

\emph{Aggregation.} Concatenating over all positions gives
$X_{\text{nn}} \in \mathbb{R}^{(RN) \times C}$. Applying a standard \textsc{Conv1D}
with $C'$ output channels, kernel size $R$, and stride $R$ yields
\begin{equation}
    y_{i,c'} = \sum_{k=1}^{R}\sum_{c=1}^{C} V[I_{i,k},\,c] \cdot w_{k,c,c'}.
    \label{eq:conv1d_agg}
\end{equation}

\emph{Recovery.} By Equation~\eqref{eq:conv1d_window}, the $k$-th selected
neighbor of an interior position is $I_{i,k} = i + (k - \lfloor R/2 \rfloor - 1)$.
Substituting and re-indexing the sum by $p = k - \lfloor R/2 \rfloor - 1$,
\begin{equation}
    y_{i,c'} = \sum_{p=-\lfloor R/2\rfloor}^{\lfloor R/2\rfloor}
               \sum_{c=1}^{C} X_{i+p,\,c} \cdot W_{p,c,c'}
             = \operatorname{Conv}(W, X)_{(i,c')},
    \label{eq:conv1d_recover}
\end{equation}
which is exactly standard 1D convolution~\eqref{eq:std_conv_1d}. Applying instead
a depthwise \textsc{Conv1D} with per-channel weights $w_{k,c}$ gives
\begin{equation}
    y_{i,c} = \sum_{k=1}^{R} V[I_{i,k},\,c] \cdot w_{k,c}
    = \sum_{p=-\lfloor R/2\rfloor}^{\lfloor R/2\rfloor}
          W_{p,c} \cdot X_{i+p,\,c}
    = \operatorname{DepthwiseConv}(W, X)_{(i,c)},
    \label{eq:conv1d_recover_dw}
\end{equation}
recovering depthwise 1D convolution~\eqref{eq:dw_conv_1d}.

\medskip
\noindent\emph{2D construction.}
Given $\mathbf{X} \in \mathbb{R}^{H \times W \times C}$, append the normalized
coordinates $p^x_{(i,j)} = i/(H-1)$ and $p^y_{(i,j)} = j/(W-1)$ along the channel
dimension and flatten to $X_{\text{pos,flat}} \in \mathbb{R}^{HW \times (C+2)}$.
Setting $N = HW$, define
\begin{equation}
    Q = K = X_{\text{pos,flat}}[:, -2{:}] \in \mathbb{R}^{N \times 2},
    \qquad V = X_{\text{flat}} \in \mathbb{R}^{N \times C}.
    \label{eq:conv2d_proj}
\end{equation}
The similarity is the exponential of the negative squared Euclidean distance 
between coordinate pairs,
\begin{equation}
    S_{(i,j),(i',j')}
    = \exp\left(
    -\left(\frac{i-i'}{H-1}\right)^2
      -\left(\frac{j-j'}{W-1}\right)^2
      \right),
    \label{eq:conv2d_sim}
\end{equation}
which is strictly decreasing in the 2D Euclidean distance between positions.
Applying $\argkmax_k$ with $K = R^2$ to each row therefore selects the $R^2$
positions nearest in 2D space; for interior positions this is the $R \times R$
centered window
\begin{equation}
    I_{(i,j)} = \left\{(i+p,\; j+q) \;\middle|\;
    p, q \in \left\{-\left\lfloor\tfrac{R}{2}\right\rfloor, \dots,
                     \left\lfloor\tfrac{R}{2}\right\rfloor\right\}\right\}.
    \label{eq:conv2d_window}
\end{equation}
The remaining steps are identical to the 1D case. A standard \textsc{Conv1D} with
kernel size $R^2$ and stride $R^2$ under uniform weights yields
\begin{equation}
    y_{(i,j),c'} = \sum_{k=1}^{R^2}\sum_{c=1}^{C}
    V[I_{(i,j),k},\,c] \cdot w_{k,c,c'}
    =
    \sum_{p=-\lfloor R/2\rfloor}^{\lfloor R/2\rfloor}
    \sum_{q=-\lfloor R/2\rfloor}^{\lfloor R/2\rfloor}
    \sum_{c=1}^{C}
    W_{p,q,c,c'} \cdot \mathbf{X}_{i+p,\,j+q,\,c}
    = \operatorname{Conv}(W, \mathbf{X})_{(i,j,c')},
    \label{eq:conv2d_recover}
\end{equation}
recovering standard 2D convolution~\eqref{eq:std_conv_2d}, and the depthwise
variant recovers Equation~\eqref{eq:dw_conv_2d} analogously,
\begin{equation}
    y_{(i,j),c} = \sum_{k=1}^{R^2}
    V[I_{(i,j),k},\,c] \cdot w_{k,c}
    =
    \sum_{p=-\lfloor R/2\rfloor}^{\lfloor R/2\rfloor}
    \sum_{q=-\lfloor R/2\rfloor}^{\lfloor R/2\rfloor}
    W_{p,q,c} \cdot \mathbf{X}_{i+p,\,j+q,\,c}
    = \operatorname{DepthwiseConv}(W, \mathbf{X})_{(i,j,c)}.
    \label{eq:conv2d_recover_dw}
\end{equation}
This establishes the claim for all interior positions in both 1D and 2D.
\end{proof}

\begin{remark}[Kernel parity]
\label{rem:kernel_parity}
For odd $R$ a natural integer center exists and the interior window is exactly
symmetric, with $\lfloor R/2 \rfloor$ neighbors on each side, giving exact
equivalence. For even $R$ there is no integer center, and $\argkmax_k$ produces
an asymmetric split (e.g., $R/2 - 1$ neighbors on one side and $R/2$ on the
other, depending on tie-breaking), so the equivalence holds only up to a
half-pixel asymmetry.
\end{remark}

\subsection{Proof of Theorem~\ref{thm:padding_equivalence}}
\label{proof:padding_equivalence}

\restatetheorem{thm:padding_equivalence}{Exact Boundary Recovery via Zero-Padding}{%
The equivalence of Theorem~\ref{thm:conv_equivalence} holds universally across
all spatial positions, including sequence boundaries, under a zero-padding scheme
matching the standard convolutional receptive field.}

\begin{proof}
Theorem~\ref{thm:conv_equivalence} establishes equivalence at interior
positions. We first characterize the boundary behavior of unpadded \convnn,
then show that zero-padding removes the discrepancy.

\medskip
\noindent\emph{Boundary skewing (unpadded case).}
At a boundary position the symmetric centered window of
Equation~\eqref{eq:conv1d_window} cannot be realized, since fewer than
$\lfloor R/2 \rfloor$ valid positions exist on one side. As $S_{ij}$ is strictly
decreasing in $|i-j|$, $\argkmax_k$ fills the remaining slots with the
next-nearest positions on the opposite side, producing a skewed window. For the
1D case:
\begin{itemize}
    \item \textbf{Left boundary} ($i < \lfloor R/2 \rfloor$): all available left
    neighbors are selected and the remaining slots are filled from the right,
    giving a right-skewed window; at the extreme $i = 0$,
    $I_0 = \{0, 1, \dots, R-1\}$.
    \item \textbf{Right boundary} ($i > N - 1 - \lfloor R/2 \rfloor$):
    symmetrically, the window is left-skewed; at the extreme $i = N-1$,
    $I_{N-1} = \{N-R, \dots, N-2, N-1\}$.
\end{itemize}
This matches the behavior of an unpadded convolution, in which kernel taps that
would fall outside the input are instead supplied by the nearest valid position
rather than by zero. Thus the unpadded \convnn\ output differs from same-padded
convolution precisely at the positions whose window would extend beyond the
sequence. The 2D case follows along each axis independently, producing top-,
bottom-, left-, or right-skewed windows at the corresponding boundaries.

\medskip
\noindent\emph{Exact recovery via zero-padding.}
Pad the input with $\lfloor R/2 \rfloor$ zeros on each side before applying
\convnn. For a 1D input this gives
$X_{\text{pad}} \in \mathbb{R}^{(N + 2\lfloor R/2\rfloor) \times C}$, and for a
2D input
$\mathbf{X}_{\text{pad}} \in \mathbb{R}^{(H + 2\lfloor R/2\rfloor) \times (W + 2\lfloor R/2\rfloor) \times C}$;
in both cases the normalized coordinates are recomputed over the padded
dimensions. Now every query position lies at least $\lfloor R/2 \rfloor$ steps
from a boundary, so $\argkmax_k$ recovers a fully symmetric centered window at
every position,
\begin{equation}
    I_i = \left\{\, i - \left\lfloor\tfrac{R}{2}\right\rfloor,\;
                 \dots,\; i,\; \dots,\;
                 i + \left\lfloor\tfrac{R}{2}\right\rfloor \,\right\}
    \quad \text{for all } i \in \{0, \dots, N + 2\lfloor R/2\rfloor - 1\}.
    \label{eq:pad_window}
\end{equation}

\emph{Output trimming.} \convnn\ produces an output over the padded spatial
dimensions; removing the $\lfloor R/2 \rfloor$-wide border recovers a tensor of
the original spatial size $N$ (or $H \times W$). Since the centered window of
Equation~\eqref{eq:pad_window} holds at every position—including those mapping to
original boundaries—the trimmed output equals same-padded convolution at all
positions.

\emph{Conclusion.} For odd $R$, the window is symmetric everywhere and the
zero-padded \convnn\ output matches Equations~\eqref{eq:std_conv_1d}--\eqref{eq:dw_conv_2d}
exactly, with no approximation. For even $R$, the half-pixel asymmetry of
Remark~\ref{rem:kernel_parity} persists even under padding, as it stems from the
absence of an integer center rather than from boundary effects.
\end{proof}

\section{Proofs for the Connection to Attention}
\label{proof:attention_generalization}

\subsection{Proof of Theorem~\ref{thm:attention_generalization}}
\label{proof:attention_generalization_proof}

\restatetheorem{thm:attention_generalization}{Generalization of Attention Frameworks}{%
Let $Q, K, V$ be projections of an input $X$ via learned linear maps, and let the
similarity matrix $S$ be the scaled dot product. Then \textup{(1)} standard
self-attention is recovered when neighbor selection is global ($K = N$) and the
depthwise aggregation weights are unit-valued ($w_{k,m} = 1$); \textup{(2)}
KVT-Attention is recovered when $K < N$ with unit aggregation weights; and
\textup{(3)} structured sparse attention (e.g., BigBird, Sparse Transformer) is
recovered by injecting a mask $M \in \{0, -\infty\}^{N \times N}$ into the
similarity, $S' = S + M$.}

\begin{proof}
We prove the three claims in turn. Throughout, $\mathbf{s}_i = \kmax_k(S)[i,:]$
and $I_i = \argkmax_k(S)[i,:]$ denote the top-$K$ similarity values and indices
for query $i$, and $\rho = \operatorname{softmax}$.

\medskip
\noindent\emph{Claim 1 (standard self-attention).}
With $Q = XW^Q$, $K = XW^K$, $V = XW^V$ and scaled dot-product similarity
$S = QK^\top/\sqrt{d_k}$, the modulated neighborhood for query $\mathbf{x}_i$ is
\begin{equation}
    X_{\text{nn},i}
    = \operatorname{diag}\!\big(\operatorname{softmax}(\mathbf{s}_i)\big)\,
      V[I_i, :] \in \mathbb{R}^{K \times d_k}.
    \label{eq:attn_nn}
\end{equation}
Aggregating with a depthwise \textsc{Conv1D} of kernel size $K$ and stride $K$,
and expanding elementwise,
\begin{equation}
    y_{i,m} = \sum_{k=1}^{K}
    \operatorname{softmax}(\mathbf{s}_i)_k \cdot V[I_{i,k},\, m] \cdot w_{k,m},
    \label{eq:attn_agg}
\end{equation}
where $w_{k,m}$ is the $k$-th kernel weight for output channel $m$. Setting
$K = N$ makes neighbor selection global: $I_i$ enumerates all $N$ positions, and
$\operatorname{softmax}(\mathbf{s}_i)$ taken over the $K=N$ scores coincides with
the full-sequence softmax of Equation~\eqref{eq:self_attention_partition}. Fixing
$w_{k,m} = 1$ then gives
\begin{equation}
    y_{i,m} = \sum_{k=1}^{N}
    \operatorname{softmax}(\mathbf{s}_i)_k \cdot V[I_{i,k},\, m].
    \label{eq:attn_recover}
\end{equation}
Writing $j = I_{i,k}$, so that the sum over neighbors $k$ is a sum over all
positions $j$, and identifying $a_{i,j} = \operatorname{softmax}(\mathbf{s}_i)_k$
and $v_j = V[j,:]$, Equation~\eqref{eq:attn_recover} is exactly
$y_i = \sum_{j=1}^{N} a_{i,j}\, v_j$ from
Equation~\eqref{eq:self_attention_partition}. Standard self-attention is
therefore recovered.

\medskip
\noindent\emph{Claim 2 (KVT-Attention).}
In KVT~\cite{DBLP:conf/eccv/WangWWLCLJ22}, for query $q_i$ the $K$ most similar
keys form a sparse neighbor set $\mathcal{N}_i^k \subset \{k_1, \dots, k_N\}$
with corresponding value set $\mathcal{N}_i^v$, and the output is
\begin{equation}
    A_i = \operatorname{softmax}\!\left(
    \frac{\langle q_i,\; (k_{j_1}, \dots, k_{j_K}) \rangle}{\sqrt{d_k}}
    \right),
    \qquad
    y_i = \sum_{l=1}^{K} A_{i,l} \cdot v_{j_l},
    \quad k_{j_l} \in \mathcal{N}_i^k,\; v_{j_l} \in \mathcal{N}_i^v.
    \label{eq:kvt}
\end{equation}
The \convnn\ aggregation of Equation~\eqref{eq:attn_agg} with $K < N$ and unit
weights $w_{k,m} = 1$ is
\begin{equation}
    y_{i,m} = \sum_{k=1}^{K}
    \operatorname{softmax}(\mathbf{s}_i)_k \cdot V[I_{i,k},\, m],
    \label{eq:kvt_recover}
\end{equation}
where $I_{i,k}$ indexes the $k$-th nearest neighbor of $\mathbf{x}_i$ under the
scaled dot product. Identifying $A_{i,l} = \operatorname{softmax}(\mathbf{s}_i)_l$
and $v_{j_l} = V[I_{i,l}, :]$ makes Equations~\eqref{eq:kvt}
and~\eqref{eq:kvt_recover} identical. Hence KVT-Attention is the special case of
\convnn\ with $K < N$ and unit aggregation weights, and together with Claim 1
this shows that \convnn\ subsumes both dense and sparse $k$-NN attention.

\medskip
\noindent\emph{Claim 3 (structured sparse attention).}
Given a mask $M \in \{0, -\infty\}^{N \times N}$ encoding the permitted pattern,
form $S' = S + M$. Since $\exp(-\infty) = 0$, the softmax assigns zero weight to
every masked entry, and $\argkmax_k(S')$ selects only among unmasked positions,
restricting neighbor selection to the pattern encoded by $M$. Aggregation then
proceeds exactly as in Claim 1, the only difference being that the softmax
normalizes over the permitted subset rather than the full sequence. BigBird~\cite{zaheer2020big}
is recovered by taking $M$ as the union of a local sliding-window pattern, a set
of global-token connections, and a random pattern; the Sparse
Transformer~\cite{child2019generating} is recovered by taking $M$ as its strided,
fixed-factorized pattern.
\end{proof}

\begin{corollary}[Causal self-attention]
\label{cor:causal}
Causal self-attention~\eqref{eq:causal_attention_matrix} is the special case of
Claim 3 obtained by setting $M_{ij} = -\infty$ for $j > i$ and $M_{ij} = 0$
otherwise, which restricts each query to its current and preceding positions.
\end{corollary}

\subsection{Proof of Theorem~\ref{thm:multihead_equivalence}}
\label{proof:multihead_equivalence}

\restatetheorem{thm:multihead_equivalence}{Equivalence under Multi-Head Topology}{%
Folding the head dimension into the batch dimension prior to neighbor selection,
a single \convnn\ operation over $\mathbb{R}^{BH \times N \times d_k/H}$ is
algebraically equivalent to running $H$ independent attention heads in parallel.}

\begin{proof}
Let $X \in \mathbb{R}^{B \times N \times d_{\text{model}}}$ be the input. Compute
$Q, K, V \in \mathbb{R}^{B \times N \times d_k}$ and partition the channel
dimension across $H$ heads, giving
$Q_H, K_H, V_H \in \mathbb{R}^{B \times H \times N \times d_k/H}$.

\medskip
\noindent\emph{Head--batch folding.}
Fold the head dimension into the batch dimension to obtain
$\tilde{Q}, \tilde{K}, \tilde{V} \in \mathbb{R}^{BH \times N \times d_k/H}$, whose
$(b,h)$-th slice is
\begin{equation}
    \tilde{Q}_{bH + h} = Q_H[b, h, :, :] \in \mathbb{R}^{N \times d_k/H},
    \label{eq:fold}
\end{equation}
and analogously for $\tilde{K}, \tilde{V}$.

\emph{Per-slice operation.} Apply the single-head \convnn\ of
Section~\ref{sec:core_framework} to $(\tilde{Q}, \tilde{K}, \tilde{V})$. For each
slice $(b,h)$ the similarity is
$S^{(b,h)} = \tilde{Q}_{bH+h}\, \tilde{K}_{bH+h}^\top \in \mathbb{R}^{N \times N}$,
and neighbor selection and aggregation produce an output
$\tilde{Y} \in \mathbb{R}^{BH \times N \times d_k/H}$.

\emph{Independence across heads.} The similarity computation, $\argkmax_k$, and
\textsc{Conv1D} aggregation all act independently along the leading (batch)
dimension, so no information flows between distinct slices. The output of slice
$(b,h)$ therefore equals that of an independent \convnn\ applied to
$Q_H[b,h], K_H[b,h], V_H[b,h]$, i.e.\ the $h$-th head for the $b$-th sample.

\emph{Recovery.} Unfolding $\tilde{Y}$ gives
$Y_H \in \mathbb{R}^{B \times H \times N \times d_k/H}$; concatenating along the
head dimension and projecting by $W^O \in \mathbb{R}^{d_k \times d_{\text{model}}}$
yields $Y \in \mathbb{R}^{B \times N \times d_{\text{model}}}$, identical to the
multi-head output of Equation~\eqref{eq:multihead_attention}. Hence one \convnn\
operation on the folded tensor equals $H$ parallel single-head \convnn\
operations, since all inter-head interactions are absent by construction.
\end{proof}

\section{Triton Kernel Design}
\label{sec:triton}
The fused kernel exploits the fact that the gather, weighting, and convolution
steps of \convnn\ aggregation are all reductions over the $K$ neighbor dimension,
and can therefore be expressed as a single loop over $K$ that accumulates into an
output vector. For each query position $i \in \{1, \dots, N\}$ in each
batch (or batch-head) slice $b \in \{1, \dots, B\}$, the kernel computes
\begin{equation}
    y_{b, i, c} = \sum_{k=1}^{K}
    \rho(\mathbf{s}_{b, i})_k \cdot
    V_b[I_{b, i, k}, c] \cdot
    W_{c,k},
    \label{eq:triton_dw}
\end{equation}
where $I_{b,i,k}$ is the $k$-th nearest neighbor of position $i$,
$\rho(\mathbf{s}_{b,i})_k$ is the corresponding softmax weight, and $W_{c,k}$ is
the depthwise convolution weight for channel $c$ and kernel position $k$. For
standard convolution, the per-channel weight $W \in \mathbb{R}^{C \times K}$ is
replaced by a cross-channel weight $\mathbf{W} \in \mathbb{R}^{C' \times C \times K}$
and the sum additionally reduces over the input channel dimension,
\begin{equation}
    y_{b,i,c'} = \sum_{k=1}^{K} \sum_{c=1}^{C}
    \rho(\mathbf{s}_{b,i})_k \cdot
    V_b[I_{b, i, k}, c] \cdot
    \mathbf{W}_{c',c,k}.
    \label{eq:triton_std}
\end{equation}

The kernel is launched on a 1D grid of $B \times N$ program instances, where $B$
is the regular batch or $B \cdot H$ for the merged batch-head of the multi-head
variant. Each instance handles all $C$ channels for a single $(b, i)$ pair via a
vectorized load--accumulate loop over $K$, using a channel block size
$\texttt{BLOCK\_C} = 2^{\lceil \log_2 C \rceil}$ to align memory accesses to
powers of two. The accumulator is maintained in \texttt{float32} regardless of
input precision and cast back to the input type (\texttt{fp16}, \texttt{bf16}, or
\texttt{fp32}) before the store, ensuring numerical stability under automatic
mixed-precision training.

Custom backward passes for both variants are implemented with
\texttt{torch.autograd.Function}. The backward pass receives the output gradient
$\nabla \mathbf{Y} \in \mathbb{R}^{B \times N \times C}$ and computes gradients
with respect to $V$, $\rho(\mathbf{s})$ (the softmax weights), and $W$ (the
convolution weight). The neighbor indices $I$ are discrete and selected by
$\argkmax$, so their gradient is \texttt{None}; gradients are scattered back to
the original $V$ positions via \texttt{scatter\_add\_}, which correctly
accumulates the case where one position in $V$ is selected as a neighbor by
multiple queries. For the standard-convolution backward, the three gradient terms
are computed with \texttt{einsum}, replacing the explicit loop over $K$ with three
batched matrix contractions over the $b, i, k, c, c'$ indices.

The complete PyTorch and Triton source code for these fused kernels is provided as a separate standalone file alongside this supplementary document.

\end{document}